\pdfoutput=1

\documentclass[11pt]{article}
\usepackage{times,latexsym}
\usepackage{url}
\usepackage{siunitx}

\usepackage[]{ACL2023}

\usepackage{times}
\usepackage{latexsym}
\usepackage{booktabs}
\usepackage{multirow}

\usepackage{fontawesome}
\definecolor{softBlue}{RGB}{30, 90, 200}

\usepackage{graphicx}
\usepackage{booktabs}
\usepackage{multirow}
\usepackage{microtype}

\usepackage[english]{babel} 
\usepackage[LGR,T1]{fontenc}
\makeatletter
\newcommand{\textgreek}[1]{%
  {\fontencoding{LGR}\fontfamily{cmr}\selectfont #1}%
}
\makeatother

\usepackage{float}
\usepackage{subcaption}
\usepackage{pifont}
\usepackage[table]{xcolor}
\usepackage{colortbl}
\definecolor{lightgray}{gray}{0.9}

\newcommand{\cmark}{\textcolor{green!60!black}{\ding{51}}}
\newcommand{\xmark}{\textcolor{red}{\ding{55}}}

\usepackage[most]{tcolorbox}
\usepackage{xcolor}
\usepackage{enumitem}
\usepackage{lipsum}
\usepackage{amsmath}
\usepackage{amssymb}

\definecolor{boxgray}{gray}{0.95}
\definecolor{sectionblue}{RGB}{0,102,204}

\usepackage{graphicx}

\usepackage{inconsolata}
\newcommand{\bs}{\textcolor{black}}

\usepackage{soul}
 \usepackage{hyperref}
 
\usepackage{xcolor}

\definecolor{neutralcolor}{RGB}{200, 220, 255}
\newcommand{\neutral}[1]{\sethlcolor{neutralcolor}\hl{#1}}

\definecolor{neutralcolor-alt1}{RGB}{204, 255, 204} 
\definecolor{neutralcolor-alt2}{RGB}{255, 204, 204}   
\newcommand{\neutralgreen}[1]{\sethlcolor{neutralcolor-alt1}\hl{#1}}     \newcommand{\neutralpink}[1]{\sethlcolor{neutralcolor-alt2}\hl{#1}}     

\hypersetup{
    colorlinks=true,
    linkcolor=darkblue,
    filecolor=darkblue,      
    urlcolor=darkblue,
    pdftitle={Overleaf Example},
    pdfpagemode=FullScreen,
    }

\usepackage{graphicx}

\usepackage{xspace}

\usepackage{ifthen}
\usepackage[commandnameprefix=ifneeded,final, todonotes={textsize=small}]{changes}

\definechangesauthor[color=magenta]{ga}
\definechangesauthor[color=olive]{jh}

\newcommand{\ga}[1]{\added[id=ga]{#1}}

\newcommand{\repo}{\url{https://github.com/g8a9/mgente-gap}}

\newcommand{\model}[1]{%
  \ifthenelse{\equal{#1}{llama8b}}{%
    \textsc{Llama 3.1 8B}\xspace
  }{%
    \ifthenelse{\equal{#1}{qwen72}}{%
      \textsc{Qwen 2.5 72B}\xspace
    }{%
      \ifthenelse{\equal{#1}{tower}}{%
        \textsc{TowerInstruct Mistral v2 7B}\xspace
      }{%
        \ifthenelse{\equal{#1}{eurollm}}{%
          \textsc{EuroLLM 9B}\xspace
        }{%
          \ifthenelse{\equal{#1}{gemma9b}}{%
            \textsc{Gemma 2 9B}\xspace
          }{%
            \ifthenelse{\equal{#1}{phi4}}{%
              \textsc{Phi 4 14B}\xspace
            }{%
              \ifthenelse{\equal{#1}{llama70b}}{%
                \textsc{Llama 3.3 70B}\xspace
              }{%
                    \ga{\textsc{Unknown Model}\xspace}
              }%
            }%
          }%
        }%
      }%
    }%
  }%
}

\newcommand{\chipbox}[2][sectionblue]{%
  \tikz[baseline=(text.base)]{%
    \node[
      rounded corners=3pt,
      fill=#1,
      text=white,
      inner sep=3.5pt,
      font=\sffamily\small
    ] (text) {#2};%
  }%
  \xspace%
}

\title{Mind the Inclusivity Gap: \\
Multilingual Gender-Neutral Translation Evaluation with \textsc{mGeNTE}}

\newcommand{\fbk}{$^{\diamondsuit}$}
\newcommand{\hamburg}{$^{\star}$}
\newcommand{\dit}{$^{\circ}$}
\newcommand{\unitn}{$^{\triangle}$}
\newcommand{\IT}{$^{\spadesuit}$}
\newcommand{\ghent}{$^{\bullet}$}

\author{Beatrice Savoldi\fbk, Giuseppe Attanasio\IT, 
\textbf{Eleonora Cupin\dit,  Eleni Gkovedarou\ghent}, \\
\textbf{Janiça Hackenbuchner\ghent},  
\textbf{Anne Lauscher}\hamburg, 
\textbf{Matteo Negri}\fbk, \\ \textbf{Andrea Piergentili}\fbk\unitn, 
\textbf{Manjinder Thind}\dit, \textbf{Luisa Bentivogli}\fbk \vspace{0.2cm} \\
  \fbk Fondazione Bruno Kessler, 
  \IT Instituto de Telecomunicações,
 \dit DIT Forlì - University of Bologna \\
 \hamburg Data Science Group - University of Hamburg, 
 \unitn University of Trento,
 \ghent LT\textsuperscript{3} - Ghent University \\
 \texttt{\href{mailto:bsavoldi@fbk.eu}{bsavoldi@fbk.eu}}}

\begin{document}
\maketitle
\begin{abstract}
Avoiding the propagation of undue (binary) gender inferences and default masculine language remains a key challenge towards
inclusive multilingual technologies, particularly when translating into languages with extensive gendered morphology. 
Gender-neutral translation (GNT) represents a linguistic strategy towards fairer
communication across languages. 
However, research on GNT is limited to a few resources and language pairs.
To address this gap, we introduce \textsc{mGeNTE}, an expert-curated resource,
and use it to conduct the first systematic multilingual evaluation of inclusive translation with state-of-the-art instruction-following language models (LMs). Experiments on en-es/de/it/el
reveal that while models can \textit{recognize} when neutrality is appropriate, they cannot consistently  \textit{produce} neutral translations, limiting their 
usability. 
To probe this behavior, we enrich our evaluation with interpretability analyses that identify task-relevant features and offer initial insights into the internal dynamics of LM-based GNT.

\noindent \raisebox{-0.5em}{\includegraphics[height=1.5em]{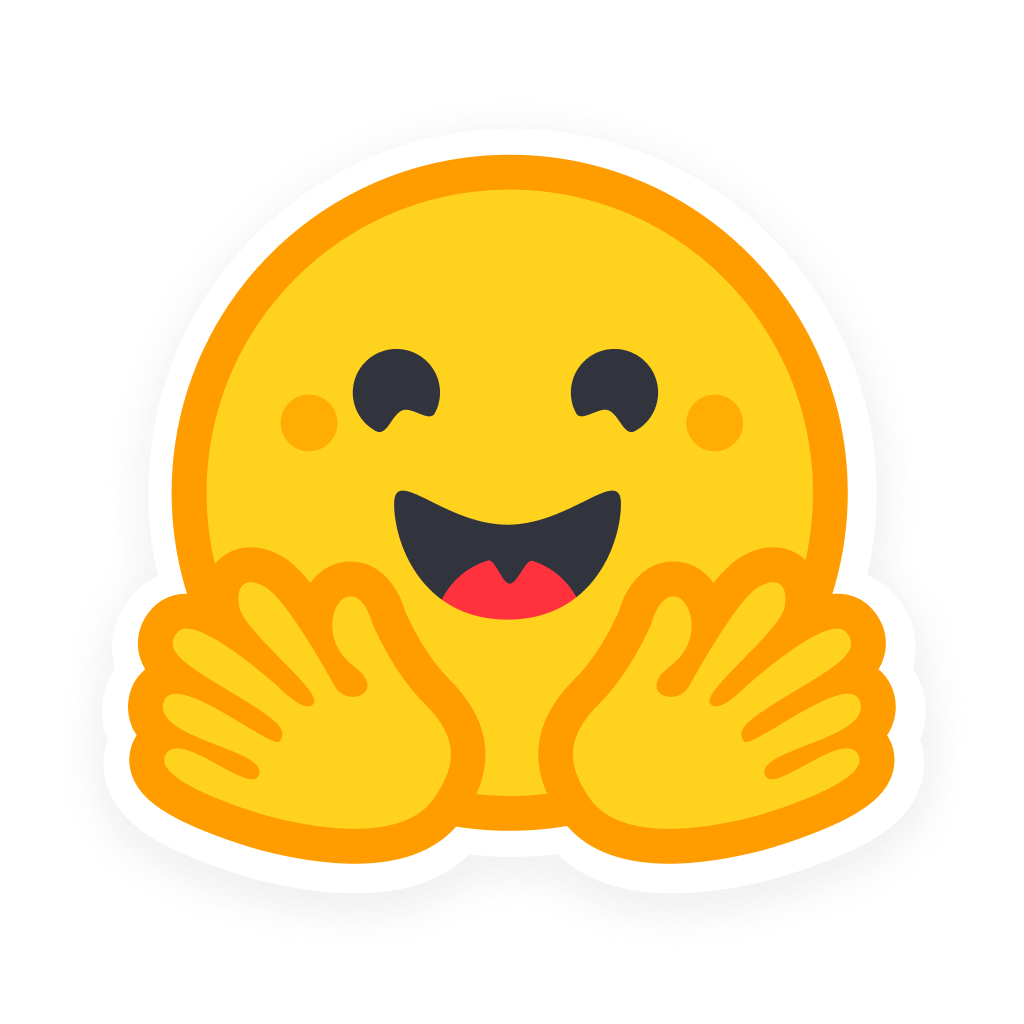}} \href{https://huggingface.co/datasets/FBK-MT/mGeNTE}{datasets/FBK-MT/mGeNTE}

\end{abstract}

\section{Introduction}
Amid societal and linguistic shifts, inclusive language practices that promote gender equality have gained relevance and use 
\citep{apa2020publication, Ashwell2023, SilvaSoares2024}.
Gender-neutral language---as an inclusive strategy also endorsed by international institutions\footnote{e.g., see the EU Parliament guidelines \url{https://www.europarl.europa.eu/cmsdata/151780/GNL_Guidelines_EN.pdf}}---advances this goal by substituting unnecessary gendered terms with unmarked forms that embrace all gender identities (e.g. \textit{spokesperson} instead of \textit{spokes\textbf{man}}) \citep{silveira1980generic, hoglund2023gendering}. 

\begin{figure}[t]
    \centering
    \includegraphics[width=1\linewidth]{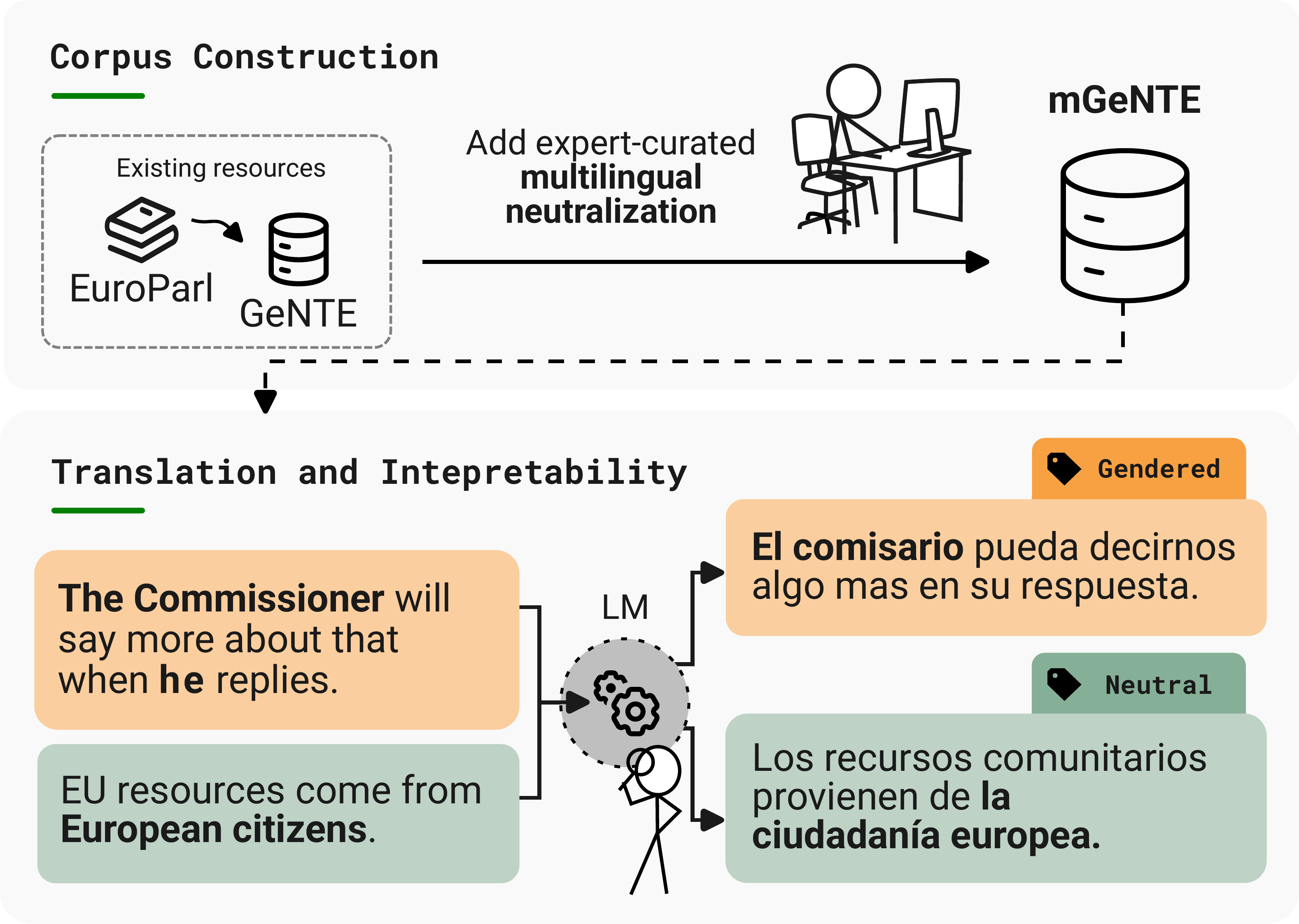}
    \caption{\textbf{Study overview} (\bs{English$\rightarrow$Spanish} example). We create a multilingual benchmark for inclusive MT \bs{in English$\rightarrow$German, Greek, Italian, and Spanish}. We test \bs{instruction-following} language models on \textit{recognizing} 
    gendered (top) vs. ambiguous source sentences (bottom) and their ability to \textit{produce} correctly gendered and neutral translations. Next, we explain models' behavior with interpretability tools. 
    }
    \label{fig:example}
\end{figure}

\begin{table*}[t]
\footnotesize
\centering
\rowcolors{2}{lightgray}{white}
\begin{tabular}{llrlccc}
\toprule
\textbf{Benchmark} & \textbf{Languages} & \textbf{Size} & \textbf{Data Type} & \textbf{Ref (Contrastive)} & \textbf{Metric} \\
\midrule
\textbf{mGeNTE (ours)}               & en-it,es,de,el & 6,000 & natural        & \cmark & \cmark \\
GeNTE  \citep{piergentili-etal-2023-hi}              & en-it              & 1,500 & natural        & \cmark & \cmark \\
Building Bridges \citep{lardelli-etal-2024-building} & en-de      & 758   & natural        & \xmark & \xmark \\
Neo-GATE  \citep{piergentili-etal-2024-enhancing}           & en-it              & 841   & naturalistic   &  \cmark & \cmark \\
GenderQueer \citep{friidhriksdottir-2024-genderqueer} & en-is             & 331   & naturalistic   &  \cmark & \cmark \\
INES  \citep{savoldi-etal-2023-test}               & de-en              & 162   & naturalistic     & \xmark & \cmark \\
Fair Translate  \citep{jourdan2025fairtranslate}     & en-fr              & 2,418 & template        & \cmark & \xmark \\
\bottomrule
\end{tabular}
\caption{Summary of multilingual resources for gender inclusive MT. We distinguish \textit{natural} data---spontaneously occurring language in authentic contexts---and \textit{naturalistic} data---human-written examples  intentionally produced to represent gender-related phenomena--and \textit{template}-based data.}
\label{tab:gender-inclusive-resources}
\end{table*}


Towards fairer technologies, prior work in natural language processing (NLP) has explored the use of inclusive language in many tasks  \citep[\textit{among others}]{sun2021they, hossain-etal-2023-misgendered,ovalle2023-towards-centering, bartl2025adapting}---%
including machine translation (MT) \citep{saunders-etal-2020-neural,
lardelli-etal-2024-sparks}. 
Indeed, translation into grammatical gender languages presents challenges due to extensive gender marking
for human referents (e.g. \textit{the citizens}$\rightarrow$ es: \textit{l\textbf{os} ciudadan\textbf{os}} \texttt{M}. vs  \textit{l\textbf{as} ciudadan\textbf{as}} \texttt{F}.) \citep{Gygaxindex4}. 
Ambiguous input often leads MT systems to default to masculine forms, reinforcing stereotypes and 
marginalizing minority gender groups 
\citep{SAVOLDI2025101257}. 

Among other efforts to mitigate exclusionary language in cross-lingual settings \citep{lauscher-etal-2023-em, daems-2023-gender}, gender-neutral translation (GNT) represents a desirable direction, also to avoid undue binary gender inferences \citep{piergentili2023gender}.
%
%
\bs{In fact, as shown in Figure~\ref{fig:example}, while preserving gendered forms in the target is justified when they are explicitly marked in the source (orange boxes), GNT offers an \textit{appropriate} alternative when gender in the source is unknown or irrelevant (green boxes).}
However, progress towards inclusive communication across languages is limited by few existing resources, and by evidence that  MT models do not support GNT \citep{savoldi-etal-2024-prompt}.


We address these limitations by introducing \textsc{\textbf{mGeNTE}}, a multilingual GNT benchmark, created by extending the language coverage and annotation layers of the existing bilingual GeNTE test set \citep{piergentili-etal-2023-hi}.
With \textsc{mGeNTE}, we examine the potential of instruction-following language models (LMs) in this space and ask: \textit{\textbf{Can multilingual LMs enable automatic neutral translation, and crucially, only when appropriate?}} 
We investigate what factors influence their ability to generate neutral outputs, evaluating five open-weight models of different sizes and families, across four configurations and four language pairs, totaling 80 conditions.


\paragraph{Contributions} 
We lay 
new groundwork for research on neutral MT with two main contributions. \textbf{(1)} A high-quality, natural resource 
covering English$\rightarrow$Italian/Spanish/German/Greek, freely available at \url{https://huggingface.co/datasets/FBK-MT/mGeNTE}.
\textbf{(2)}~The first systematic multilingual evaluation of open LMs for GNT. 
To this end, we pair downstream performance measures with interpretability analyses that shed light on the internal mechanisms supporting LM-based GNT. We release code and artifacts  at \repo.

\paragraph{Findings}~~We show that---while models \textit{recognize} when neutrality is appropriate across all language pairs---they do not consistently \textit{produce} neutral outputs (\S\ref{sec:gnt_results}). Model size and prompt context impact translation outcomes, with larger models better at leveraging context information to improve GNT.
The distinction between \textit{recognizing} and \textit{producing} is supported by an interpretability analysis (\S\ref{sec:context}), showing how such tasks rely on different context signals.  Our findings inform the current limits in usability of LMs for GNT and point to better integration of these two steps as a potential area for future improvement in inclusive MT.

\section{Background}

With a growing awareness that language encodes social inequalities and can negatively affect perceptions of gender and identity \citep{stahlberg2007representation, sczesny2016can}, research on gender inclusive language technologies has been gaining traction. In monolingual NLP, this has led to a bulk of studies covering different tasks \citep{bunzeck-zarriess-2024-slayqa, subramonian2025agree}, with most work centered on English \citep[\textit{inter alia}]{cao-daume-iii-2020-toward, bartl-leavy-2024-showgirls, gautam2024}, and emerging efforts in grammatical gender languages like German \citep{amrhein-etal-2023-exploiting, waldis-etal-2024-lou}, Portuguese \citep{veloso-etal-2023-rewriting}, and Italian \citep{attanasio2021mimic, frenda2024gfg, greco-emimic}.
%

In the multilingual and translation contexts, however, research remains comparatively scarce. 
Earlier work has shown that MT systems struggle to preserve neutrality \citep{cho-etal-2019measuring}, especially when translating into grammatical gender languages \citep{saunders-byrne-2020-addressing, piergentili-etal-2023-hi, lardelli-etal-2024-building}. 
Recently, 
LMs have been explored to meet the demands of automatic inclusive translation.  
Yet, the current landscape is fragmented. Individual studies  focus on different language pairs, linguistic phenomena, and evaluation approaches, making it difficult to compare findings.
%
Notably, \citet{savoldi-etal-2024-prompt} evaluate GNT for en-it with one \textit{closed} model only,  and exclusively on test instances that always require neutralizations---leaving unexplored whether LMs can discern when neutrality is desirable. 
\citet{jourdan2025fairtranslate}
investigate open LMs capabilities in translating both gendered forms and neomorphemes\footnote{i.e., innovative linguistic solutions. See also \S\ref{sec:ethic}.} into French---but their testbed consists of  sentences with limited variability and  
relies on standard MT metrics like 
COMET \citep{rei-etal-2021-references}, which are ill-equipped for dedicated analyses and can favor masculine forms \citep{zaranis2025watchingwatchersexposinggender}.

Overall, a current limitation is the lack of a robust, multilingual benchmark to 
address cross-lingual inclusivity concerns. As illustrated in Table~\ref{tab:gender-inclusive-resources}, existing resources are limited---in language coverage, size \citep{savoldi-etal-2023-test}, naturalness \citep{jourdan2025fairtranslate}, or lacking dedicated evaluation protocols \citep{lardelli-etal-2024-building}.  
With \textsc{mGeNTE}, we fill this gap.  Rather than building our resource from scratch, we enrich the existing GeNTE resource, which is selected for being \textit{i)} the sole natural dataset available, and \textit{ii)} allowing consistent extension to multiple language pairs to enable comparable, multilingual evaluation of GNT. Additionally, 
\bs{we complete our benchmarking effort by pairing}
\textsc{mGeNTE} with a dedicated  evaluation framework  for assessing  GNT  (\S\ref{subsec:evaluation}).

\section{The \textsc{mGeNTE} Corpus}

We create \textsc{mGeNTE} by extending the GeNTE benchmark \citep{piergentili-etal-2023-hi}—originally designed for English-Italian—to three additional language pairs: English-German, English-Spanish, and English-Greek.\footnote{\bs{These target languages were selected because, as grammatical gender languages, they have rich gendered morphology, making them suitable for GNT. In addition to three widely studied Indo-European languages from distinct branches (Italian, Spanish, German), we deliberately included Greek---a lower-resource language with a distinct script---to broaden the linguistic diversity in \textsc{mGeNTE} and to ensure the coverage of underrepresented, challenging cases in inclusive MT research.}}
Built from naturally occurring data in the Europarl corpus \citep{koehn2005europarl}, \textsc{mGeNTE} preserves the original design rationale and curation methodology to ensure comparability. Overall, each language pair comprises 1,500  <\textit{source, target-gendered}> and newly created <\textit{target-neutral}> triplets aligned at the sentence-level, resulting in a total of 6,000 entries.

As represented in Figure \ref{fig:example},
\textsc{mGeNTE} entries are evenly distributed across two translation scenarios that allow benchmarking of models’ ability to perform neutral translations, but only where appropriate:
\textit{i)}~\textsc{Set-N}, featuring gender-ambiguous human referents in the source, for which a neutral translation is desirable in the target; and
\textit{ii)} \textsc{Set-G}, featuring explicit gender mentions in the source that should be correctly rendered with gendered (masculine or feminine) forms in the target. 

To ensure \textit{high-quality}, we entrust data extraction (\S\ref{subsec:mgente-expansion}) and annotation (\S\ref{subsec:mgente-word})  to experts and translation students specialized in GNT with native or C1 competence in their assigned target language.\footnote{They are all authors of this paper.} Creating new neutral references in the target is tasked to professional translators (\S\ref{subsec:mgenter-ref}).

\bs{Final statistics of \textsc{mGeNTE} segments and annotations are given in Table \ref{tab:compact-data-stats}.}

\begin{table}[t]
\centering
\footnotesize
\begin{tabular}{lcccccc}
\toprule
& \multicolumn{2}{c}{\textbf{Segments}} & \multicolumn{2}{c}{\textbf{Gendered Words}} \\
\cmidrule(r){2-3} \cmidrule(r){4-5}
\textsc{mGeNTE} & \textbf{Set-G} & \textbf{Set-N} & \textbf{\#} & \textbf{\#Unique} \\
\midrule
\textit{en-it}  & 750 & 750 & 4115 & 802 \\
\textit{en-es}  & 750 & 750 & 4363 & 644 \\
\textit{en-de}  & 750 & 750 & 3977 & 613 \\
\textit{en-el}  & 750 & 750 & 3736 & 743 \\
\midrule
\textsc{Parallel} & 578 & 409 & -- & -- \\
\bottomrule
\end{tabular}
\caption{Distribution of \textsc{mGeNTE} segments by subset and language pair,
sentences that are fully parallel across all pairs (\textsc{Parallel}),
and total (\#) and unique (\#Unique) annotated gendered words per language.}
\label{tab:compact-data-stats}
\end{table}


\begin{table*}[t]
\centering
\scriptsize
\begin{tabular}{lll}
\toprule
\textbf{\textsc{Set-N}} & \textsc{src} & \textbf{Pensioners} are in favour of strengthening criminal law, [...] \\
\midrule

\multirow{2}{*}{\textit{en-it}} 
& \textsc{Ref-G}  & \textbf{I pensionati} sono favorevoli a un rafforzamento del diritto penale, [...] \\
& \textsc{Ref-N}$_{1}$ & \neutral{\textbf{Le persone pensionate}}\textsubscript{\texttt{[pensioned people]}} sono favorevoli a un rafforzamento del diritto penale, [...] \\
[1.5mm]

\multirow{2}{*}{\textit{en-es}} 
& \textsc{Ref-G}  & \textbf{Los pensionistas} están a favor de reforzar el Derecho penal no solo nacional, [...] \\
& \textsc{Ref-N}$_{1}$ & \neutral{\textbf{Hay pensionistas}}\textsubscript{\texttt{[there are pensioners]}} que están a favor de reforzar el Derecho penal no solo nacional, [...] \\
[1.5mm]

\multirow{2}{*}{\textit{en-de}} 
& \textsc{Ref-G}  & Die \textbf{Rentner} begrüßen den Ausbau nicht nur des einzelstaatlichen, [...] \\
& \textsc{Ref-N}$_{1}$ & \neutral{\textbf{Die Menschen in Rente}}\textsubscript{\texttt{[people in retirement]}} begrüßen den Ausbau nicht nur des einzelstaatlichen, [...] \\
[1.5mm]

\textit{en-el} 
& \textsc{Ref-G} & \textgreek{Οι \textbf{συνταξιούχοι} είναι υπέρ της ενίσχυσης του ποινικού δικαίου, [...]} \\
& \textsc{Ref-N}$_{1}$ & \textgreek{\neutral{\textbf{Τα συνταξιοδοτημένα άτομα}}}\textsubscript{\texttt{[the retired individuals]}} \textgreek{είναι υπέρ της ενίσχυσης του ποινικού δικαίου, [...]} \\
\toprule
 \textbf{\textsc{Set-G}}
 & \textsc{src}  & I trust the \textbf{Commissioner} will promise that \underline{he} will exercise extra vigilance. \\ 
 \cmidrule{2-3}
\textit{en-it }& \textsc{Ref-G}  &  Spero che \textbf{il Commissario} ora prometta di vigilare attentamente a tale riguardo. \\ 
& \textsc{Ref-N}$_{1}$ & Spero che \neutral{ \textbf{il membro della Commissione}}\textsubscript{\texttt{[the member of the board]}} ora prometta di vigilare attentamente a tale riguardo. \\ 
[1.5mm]
\textit{en-es} & \textsc{Ref-G}  &  Espero que \textbf{el Comisario} prometa controlar exhaustivamente esta situación. \\ 
& \textsc{Ref-N}$_{1}$ & Espero que \neutral{\textbf{la representación de la Comisión}}\textsubscript{\texttt{[the representative of the board]}} prometa...\\ 
[1.5mm]
\textit{en-de} & \textsc{Ref-G}  & Von \textbf{dem Herrn Kommissar} erwarte ich heute die Zusage, \textbf{er} werde mit Argusaugen darüber wachen.  \\ 
& \textsc{Ref-N}$_{1}$ & \neutral{\textbf{Von dem Kommissionsmitglied}}\textsubscript{\texttt{[From the board member]}} erwarte ich heute die Zusage, \neutral{\textbf{es}}\textsubscript{\texttt{[they]}} werde mit Argusaugen... \\ 
[1.5mm]
\textit{en-el} & \textsc{Ref-G} & \textgreek{Προσδοκώ από \textbf{τον} Επίτροπο να δεσμευτεί ότι θα επιβλέψει αυστηρά την κατάσταση}. \\
& \textsc{Ref-N}$_{1}$ & \textgreek{Προσδοκώ από \textbf{\neutral{το μέλος της Επιτροπής}}}\textsubscript{\texttt{[the member of the Commission]}} \textgreek{να δεσμευτεί ότι θα επιβλέψει [...].}\\

\bottomrule
\end{tabular}
\caption{\textbf{\textsc{mGeNTE Parallel}}. Entries from \textsc{Set-N} and \textsc{Set-G}, with gendered references (\textsc{Ref-G}) and \textsc{Ref-N} with \neutral{neutralizations}. 
 Words in \textbf{bold} mention human referents; \underline{underlined} source words express the referent's gender.}
\label{tab:parallel}
\end{table*}

\subsection{Multilingual Expansion and Alignment}
\label{subsec:mgente-expansion}

\paragraph{Parallel data extraction}
We prioritize selecting sentences aligned across en-it/es/de/el. 
To this aim, we started by retrieving the Europarl\footnote{Specifically, from the \texttt{common test set 2}, which contains data aligned across multiple languages: \url{https://www.statmt.org/europarl/archives.html} 
} sentences that are already contained in GeNTE en-it.
Each automatically extracted sentence\footnote{Using the original Europarl IDs available in GeNTE.} was checked to confirm that it contained a gender-related phenomenon; if not, we discarded it.\footnote{This is possible due to cross-linguistic differences in how different languages encode and express genders. \bs{For instance, ``child'' is gender-marked in Italian (\textit{\textbf{il/la} bambin\textbf{o}/\textbf{a}}) but remains invariant in German (\textit{das Kind}).}}
This process yielded 987 sentences that are fully parallel across all four pairs, henceforth referred to as \textsc{Parallel-Set}.

\paragraph{Language-specific extraction}
To reach the target of 1,500 sentences per language pair, the remaining sentences were extracted \textit{ex novo} using regular expressions. These were chosen to represent both subsets (\textsc{Set-N} or \textsc{Set-G}) as well as language-specific gender patterns.
Accordingly, for \textsc{Set-G} we targeted unambiguous English source sentences, i.e. containing explicit gender cues such as titles (\textit{Mr, Mrs}) and  pronouns (\textit{him, her}). Vice versa, for \textsc{Set-N} we excluded segments with source gender information and matched expressions that---while gender ambiguous in English---correspond to either masculine or feminine forms in the target language (e.g., \textit{deputy} $\rightarrow$ es: \textit{deputad\textbf{o}/\textbf{a}}, de: \textit{Stellvertret\textbf{er}/\textbf{erin}}, 
el: \textgreek{\textit{αναπληρω\textbf{τής/τρια}}}).

\paragraph{Sentence Alignment}
To align with the original GeNTE corpus and streamline evaluation, source sentences containing multiple referents requiring different gender forms were edited to consistently require a single form in the target language.\footnote{In this way, each entry can be treated as a coherent gendered or neutral unit.} Additionally, to address the under-representation of unambiguous feminine data in \textsc{Set-G}, the corpus was adjusted through gender-swapping interventions to achieve a balanced distribution of feminine and masculine forms.\footnote{Masculine forms are over-represented in the original target references, reflecting a well-known under-representation of women and feminine forms in existing resources \citep{vanmassenhove-etal-2018-getting, gaido-etal-2020-breeding}.} 
Minor corrections (e.g., typos, translation errors) were also applied. Full details of the editing process and intervention statistics are provided in Appendix \ref{app:mgente}. At the end of the editing process, all sentences were manually sorted into \textsc{Set-N}, or as either feminine (F) or masculine (M) instances from \textsc{Set-G}. 
Such a distinction is central to \textsc{mGeNTE} design and explored in the following analyses.



\subsection{Gendered Words Annotation}
\label{subsec:mgente-word}
To further enrich \textsc{mGeNTE}, all target sentences---including the original en-it ones---were manually annotated at the word level
\footnote{
\bs{We compute inter-annotator agreement (IAA) 
on the exact matches of the gendered words annotated
by two annotators. IAA  (Dice coefficient) was $\geq 0.92$. For further details and a breakdown by language, see Appendix \ref{app:mgente}.}} to identify gendered words requiring neutralization, enabling the exploration of
cross-linguistic variations in gender phenomena. 
\bs{As shown in Table \ref{tab:compact-data-stats},}
the corpus displays substantial lexical variability, with over 600 unique gendered words per language. Notably, Italian and Greek exhibit a higher unique  count of gendered words, reflecting their morphosyntactic tendency to mark gender not only on nouns and adjectives but also on verbal forms.
Also, based on qualitative observations, \textsc{Set-N} annotated words are vastly 
populated with masculine forms used generically to refer to mixed or unknown referents. These cases, though translated with gendered (masculine) forms in the original Europarl references, are exactly the target of 
language neutrality
efforts 
\citep{gygax2021masculine}
and well-suited candidates for GNT. 
For more insights and details on the annotation process, see Appendix \ref{app:mgente}.


\subsection{Gender-Neutral Reference Creation}
\label{subsec:mgenter-ref}
For each German, Spanish and Greek Europarl (gendered) 
reference translation---
henceforth \textsc{Ref-G}---we created an additional gender-neutral reference (i.e. \textsc{Ref-N}), which differs from the original one only in that it refers to human entities with neutral expressions.
Neutralization is an open-ended task with a high degree of variability in its possible solutions. In our target languages with rich gendered morphology, this can involve a range of strategies \citep{ papadopoulos2022brief, ScottoDiCarlo2024, muller2024less}---from local lexical changes (e.g. epicene synonyms, collective nouns) to more extensive rephrasings (e.g. impersonal or passive constructions) to be selected contextually to preserve the adequacy 
of the neutral translations. 

We ensured high-quality, diversified and contextually appropriate neutralizations by commissioning three professional translators per language pair---selected for their expertise in neutral language---and informed by detailed language-specific neutralization guidelines.%
\footnote{The guidelines are available with the \href{https://huggingface.co/datasets/FBK-MT/mGeNTE}{data release}. Translators were compensated at market rates: €25/hr for en-es/el and €35/hr for en-de.}

\paragraph{\textsc{Common Set} }

Following the GeNTE design, we additionally created a \textsc{common} of 200 sentences (100 from each \textsc{Set}) to be neutralized by all translators. We thus obtain three  \textsc{Ref-N}s per source sentence.
The \textsc{common} set, besides adding a richer dimension to the corpus, allows us to ask:  \textit{how much variability do translators yield when neutralizing the same sentence?} Our analysis reveals considerable variation across language pairs, with identical neutralizations occurring in only 11\% (en-es), 9.3\% (en-de), and 14.9\% (en-el)\footnote{For the original en-it it amounts to 13.5\%.} of cases. This variability is directly reflected in the corpus, attesting to its natural diversity, an asset for studying and evaluating open-ended tasks like GNT.

We show a full entry from the \textsc{common} set in Table \ref{tab:mgente-common-examples} Appendix \ref{app:mgente}, and refer to Table \ref{tab:parallel} for a \textsc{Parallel} entry example across all language pairs. Generally, in ambiguous \textsc{Set-N} entries (top), translators avoid masculine generics through rephrasing strategies, whereas neutralization of gender-unambiguous sentences \textsc{Set-G} can be more verbose and unnecessary. 
\begin{figure}[t]
\begin{minipage}[t]{0.92\linewidth}
\begin{tcolorbox}[
colback=boxgray,
  colframe=gray,
  sharp corners,
  boxrule=0.8pt,
  top=6pt, bottom=6pt, left=5pt, right=5pt,
label={box:translation-prompt}
]
\begin{scriptsize}

You are a helpful \texttt{\{lang\}} translator specialized in gender-neutral language.

\tcbline

Translate the following sentences from English into \texttt{\{lang\}} following these rules:
\begin{enumerate}[left=2pt, itemsep=0.5pt, parsep=3pt, topsep=3pt]
  \item If the source English sentence clearly indicates gender for human referents (masculine or feminine): Translate using gendered language and use the label **GENDERED**
  \item If the source English sentence does not indicate gender for human referents: Translate using gender-neutral language and use the label **NEUTRAL**
\end{enumerate}

\tcbline

Guidelines for Gender-Neutral Translation:
\begin{itemize}[left=2pt, itemsep=0.5pt, parsep=0pt, topsep=0pt]
  \item Use neutral synonyms 
  \item Use neutral collective nouns 
  \item Use neutral rephrasings 
  \item Avoid masculine forms for generic referents 
  \item Avoid neomorphemes
  \item Avoid double feminine/masculine forms 
\end{itemize}

\end{scriptsize}
\tcblower
\begin{scriptsize}
    


\textbf{user:}  <en> \texttt{\{English source\}} \\
\textbf{assist.:} <\texttt{\{lang\}}> **GENDERED** [\texttt{\{gendered translation\}}]

\vspace{0.5em}
\textbf{user:}  <en> \texttt{\{English source \}} \\
\textbf{assist.:} <\texttt{\{lang\}}> **GENDERED** [\texttt{\{gendered translation\}}]


\vspace{0.5em}
\textbf{user:}  <en> \texttt{\{English source \}} \\
\textbf{assist.:} <\texttt{\{lang\}}> **NEUTRAL** [\texttt{\{neutral translation\}}]


\vspace{0.5em}
\textbf{user:}  <en> \texttt{\{English source \}} \\
\textbf{assist.:} <\texttt{\{lang\}}> **NEUTRAL** [\texttt{\{neutral translation\}}]


\end{scriptsize}
\end{tcolorbox}
\end{minipage}%
\begin{minipage}[t]{0.10\linewidth}
\begin{tikzpicture}[remember picture, overlay]
    \coordinate (system) at (0.5\linewidth, +9cm);
    \coordinate (preamble) at (0.5\linewidth, +6.5cm);
    \coordinate (guidelines) at (0.5\linewidth, +3.5cm);
    \coordinate (examples) at (0.5\linewidth, 0.8cm);
    
    \node[rotate=90, anchor=west, font=\small\bfseries\color{sectionblue}] at (system) {\texttt{Sys}};
    \node[rotate=90, anchor=west, font=\small\bfseries\color{sectionblue}] at (preamble) {\texttt{Preamble}};
    \node[rotate=90, anchor=west, font=\small\bfseries\color{sectionblue}] at (guidelines) {\texttt{Guidelines}};
    \node[rotate=90, anchor=west, font=\small\bfseries\color{sectionblue}] at (examples) {\texttt{Exemplars}};
\end{tikzpicture}
\end{minipage}

\caption{GNT prompt overview with labeled sections. The prompt consists of system instructions, translation rules (preamble), gender-neutral guidelines, and exemplars provided as conversational turns.}
\label{translation-prompt}

\end{figure}

\section{Experimental Setup}
We conceptualize the GNT task as a model’s ability to produce correctly gendered translations when the source specifies gender (i.e. \textsc{Set-G}), and neutral translations when 
the source refers
to unspecified referents (i.e. \textsc{Set-N}). In our setup---to disentangle source sentence categorization from generation performance---we enforce models to output both (\textit{i}) a  \textit{label} indicating the source category, as well as (\textit{ii}) the 
\textit{translation} of the source sentence.

\subsection{Models setup}
\label{subsec:models}

\paragraph{Models} 

We experiment with open-weight multilingual models.
Starting from an initial pool of 10 SOTA models\footnote{As of March 2024 on \href{https://huggingface.co/open-llm-leaderboard}{Open LLM Leaderboard}.} from 5 families (Qwen, LLama, Mistral, Gemma, Phi), we used translation quality and format adherence
as thresholds for inclusion in the main experiments (details in Appendix \ref{app:preliminar_results}). Based on these criteria, we retain five final instruction models of varying sizes: \model{llama8b} and \model{llama70b} \citep{grattafiori2024llama3herdmodels}, \model{qwen72} \citep{qwen2024}, \model{gemma9b} \citep{gemmateam2024gemma2improvingopen}, and \model{phi4} \citep{abdin2024phi4technicalreport}. Experimental details are in Appendix \ref{app:model_prompting}.

\begin{figure*}[t]
    \centering
    \includegraphics[width=1\linewidth]{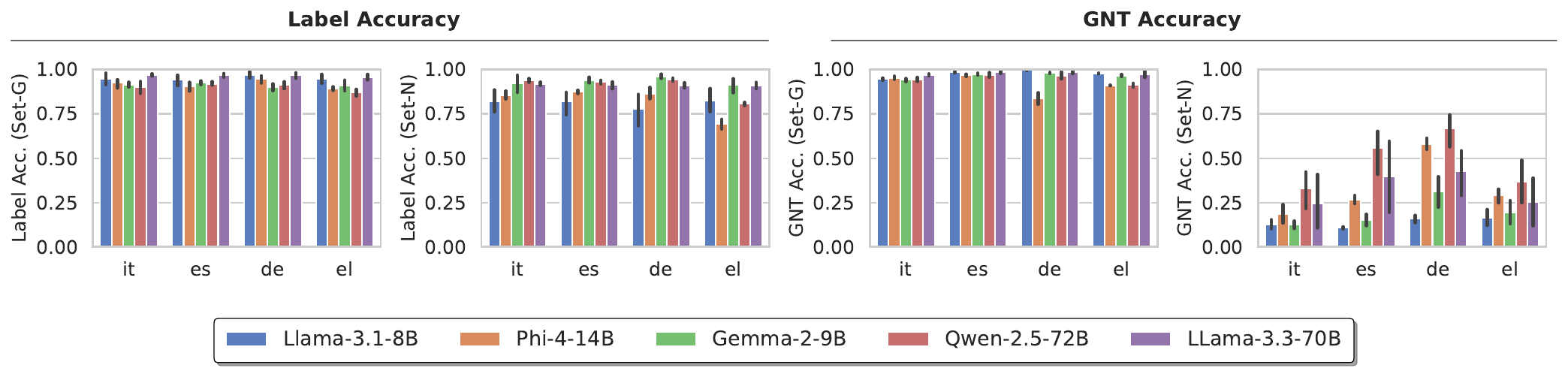}
    \caption{Source category (\textit{left}) and GNT accuracy (\textit{right}) results across \textsc{mGeNTE} Sets (avg. across prompts).}
    \label{fig:GNT-scores}
\end{figure*}

\paragraph{Setup} We test models' in-contex-learning 
capabilities \citep{brown} for the GNT task. 
%
Our prompt (shown in Figure \ref{translation-prompt}) includes a system prompt (\textit{Sys}), the task description (\textit{Preamble}), language-specific GNT \textit{guidelines},\footnote{Depending on the target, different linguistic examples are provided. Full prompts available in our \href{https://github.com/g8a9/mgente-gap}{repository}.} and four task demonstrations (2 gendered from \textsc{Set-G}, 2 neutral from \textsc{Set-N}) randomly sampled from \textsc{mGeNTE} \textsc{parallel}
and excluded from evaluation. 
%
To 
analyze 
robustness and the impact of context components, 
we test four different configurations: including both the system prompt and the guidelines, excluding one of them, or excluding both---
totaling \num4 \texttimes \num5 model configurations per language pair.\footnote{We always provide the task definition and the four shots, though our preliminary experiments in Appendix \ref{app:model_prompting} also include zero-shot and 2-shot settings.}



\subsection{Evaluation}
\label{subsec:evaluation}

\paragraph{Overall quality} We assess the selected models’ translation quality using xCOMET  \citep{guerreiro-etal-2024-xcomet},\footnote{\url{https://huggingface.co/Unbabel/XCOMET-XL}} scoring outputs against their respective correct references (\textsc{Ref-G} for \textsc{Set-G}, \textsc{Ref-N} for \textsc{Set-N}). The models achieve high average scores (en-de 0.96, en-es 0.95, en-it 0.95), with
 Greek---being a 
 less
 supported language---showing comparatively lower average performance (0.83).


\paragraph{Source sentence category} As a first inclusivity-related metric, we measure the \textit{accuracy} of label generation (either \textit{Gendered} or \textit{Neutral}) against the gold \textsc{Set-G/N} annotations in \textsc{mGeNTE}.

\paragraph{Gender-Neutral Translation}
We evaluate the \textit{accuracy} of models in producing correctly gendered and neutral translations by using an \textit{LLM-as-a-judge} approach \citep{gu2025surveyllmasajudge}, which enables scalable GNT evaluation across multiple languages.
%
In practice, we adapt the structured approach proposed by \citet{piergentili2025llmasajudgeapproachscalablegenderneutral}, which provides sentence-level neutrality binary judgments and was tested on human-written gendered vs. neutral text.\footnote{The original GeNTE evaluation method is limited to en~-it, and is superseded by the LLM-as-a-judge approach by  \citet{piergentili2025llmasajudgeapproachscalablegenderneutral}.}
We rely on their optimal prompt\footnote{i.e. Cross+P+L in the original paper.} by adapting it to all languages covered in \textsc{mGeNTE}, and validate its effectiveness on automatic translations. 
We tested different LLMs on  
1,000 manually annotated model outputs.
Our best-performing evaluation setup relies on \textsc{GPT-4o},\footnote{\bs{gpt-4o-2024-08-06}} which achieves 0.87 macro F1  and 92\% accuracy. 
Full results are in Table \ref{tab:eval_results} (Appendix \ref{app:llm-judge}), along with prompt and data annotation details.

\section{Gender-Neutrality Results}
\label{sec:gnt_results}
We present GNT results on the \textsc{mGeNTE} benchmark. Full multilingual results for both label and translation generation are in Figure~\ref{fig:GNT-scores}.



\paragraph{LMs effectively distinguish ambiguous from gendered source sentences.} Source category scores in Figure~\ref{fig:GNT-scores} show strong, consistent performance across languages, models, and with minor variance in prompt configurations (see error bars). Overall, \textit{gendered} accuracy is only slightly higher.\footnote{Worst overall scores are by \model{phi4} for el Set-N $<75$. Complete results are available in Table~\ref{tab:macro_f1_lang_model_2dp} in Appendix~\ref{app:results}.}



%

\paragraph{GNT is challenging, with variations across models and languages.}
Figure~\ref{fig:GNT-scores} shows that while \textsc{Set-G} sentences are consistently translated with the correct gendered forms, accuracy on \textsc{Set-N} is systematically lower and characterized by higher variance (see error bars), with difficulty in producing neutral translations. Results vary by language: en-el/it achieves the lowest rates of correct GNTs. Greek’s results align with its lower overall generic performance (\S\ref{subsec:evaluation}), but en-it’s underperformance is more surprising given its otherwise solid overall translation quality. Broadly, larger models outperform smaller ones, with \model{qwen72} leading overall, followed by \model{llama70b} (en-es, en-it) and \model{phi4} (en-de, en-el).

\begin{figure}[t]
    \centering
    \includegraphics[width=1\linewidth]{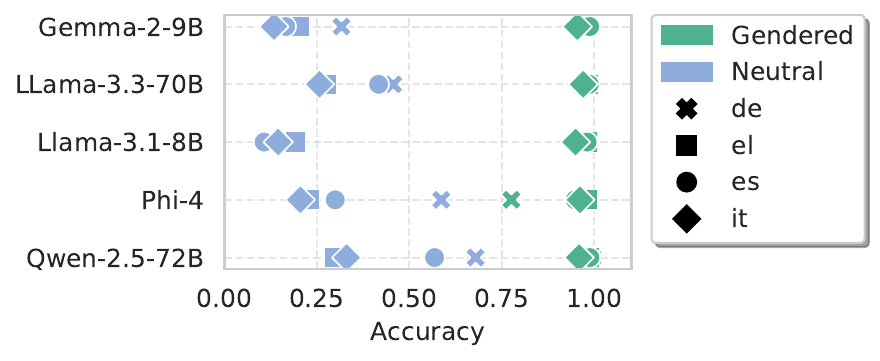}
    \caption{\textbf{Accuracy of label-translation coherence}. Measures the agreement of translation forms (gendered/neutral)  with the generated label. Scores are averaged across prompt configurations.
    }
    \label{fig:gnt-labell}
\end{figure}

\paragraph{Correct source categorization does not guarantee correct GNT.}
Given the mismatch between source categorization and GNT performance, in Figure~\ref{fig:gnt-labell} we measure how consistently the output translation form (gendered/neutral) and the generated label are in agreement. These results highlight a  gap between recognizing the source sentence category and 
generating 
the appropriate
translation. 
Models systematically produce gendered translations when assigning a gendered label,
but coherence drops sharply for neutral cases---often 
below random chance. Only \model{qwen72} and \model{phi4} maintain 
higher agreement in their strongest language pairs (en-es/de and en-de, respectively). 

\paragraph{(Larger) LMs benefit from richer context.}

\begin{figure}[t]
    \centering
    \includegraphics[width=1\linewidth]{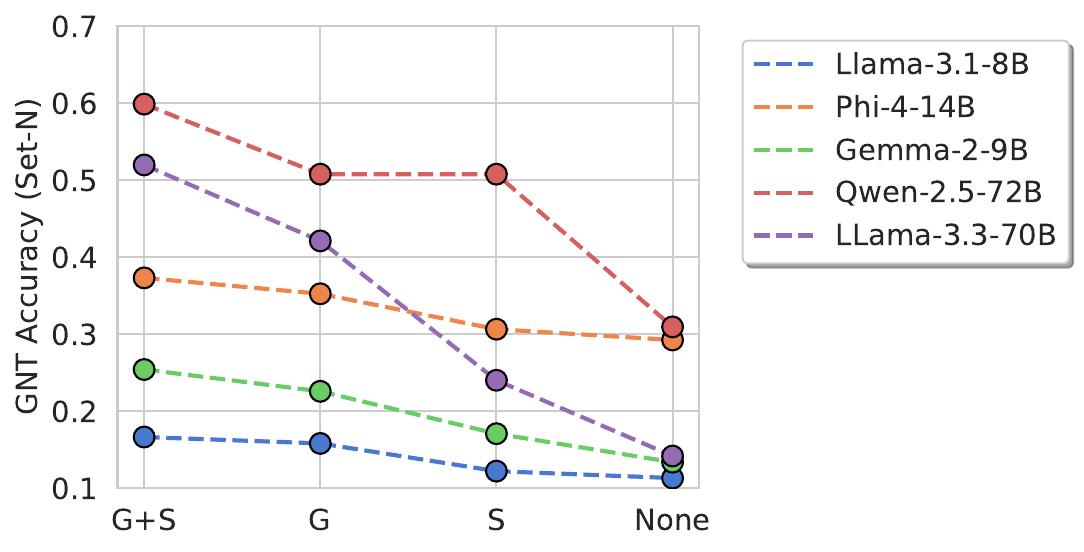}
    \caption{\textbf{GNT Accuracy (Set-N) across prompt configurations}, i.e. using both context parts \textit{system} and \textit{guidelines} (G+S), only one of them, or \textit{None}. Markers show values averaged over all language pairs.}
    \label{fig:prompt-config}
\end{figure}

Figure~\ref{fig:prompt-config} breaks down GNT results by prompt configuration  to explore the impact of provided context.\footnote{Focusing on \textsc{Set-N} GNT. Full disaggregated results across Sets are in Appendix \ref{suapp:GNT-results}.} 
Including both \textit{guidelines} and \textit{system} (G+S)  yields higher GNT accuracy, while removing both (\textit{None}) 
leads to the worst performance.
Larger models 
show higher gains from 
rich
prompts, 
better 
leveraging in-context information. Notably, while \model{qwen72} maintains relatively higher GNT performance across setups, \model{llama70b} drops to small-model levels in the \textit{None} setup. For individual language pairs results, see Figure \ref{fig:config-langs} in Appendix  \ref{suapp:GNT-results}. 

\paragraph{{\color{softBlue}\faSearch}} Overall, our findings highlight that (\textit{i}) though LMs reliably detect when gender neutrality is needed, they do not consistently produce neutral translations, with GNT capabilities notably less robust. 
GNT (\textit{ii}) varies across languages;\footnote{Potentially reflecting both data availability and sociolinguistic factors---the higher performance for en-de may relate to greater distribution and progress in language inclusivity for German; we leave this for future research.} (\textit{iii}) depends on model choice---with larger models generally performing better---and  (\textit{iv}) is  sensitive to prompt variations---indicating that contextual information in prompts can affect real-world usability of LMs towards inclusive translation. 

To better understand this behavior, we use 
for the first time
context attribution and explainability techniques to shed light on how LMs handle gender-neutral translation.

\section{Context Analysis}
\label{sec:context}
Section~\ref{sec:gnt_results} established that source category detection and GNT translation differ in performance, and the prompt has a decisive impact on GNT. Hence, it is of utmost interest to measure \textit{when} and \textit{how} models use contextual information when carrying out GNT. 
This can help explain observed mismatches, guide future development, and support immediate use of LMs by enabling prompt steering for inclusivity.

Inspired by recent long-context attribution work \citep{sarti2024quantifying,cohen2024contextcite,liu2024attribot}, we hence resolve to post-hoc interpretability \bs{\citep{madsen2022post}} for a finer-grained analysis. 
We are interested in assessing the contribution of each input token to generating a specific output. This type of task is commonly known as \textit{feature attribution}, 
and several methods have been proposed for NLP task \citep[][among others]{mosca-etal-2022-shap,ferrando-etal-2022-towards}, including MT \citep{zaranis-etal-2024-analyzing}. Prior work has leveraged interpretability for gender bias in MT \citep{sarti-etal-2023-inseq, attanasio-etal-2023-tale}, focusing on binary gender and source token (i.e., pronouns) contributions. Our study extends them to multilingual GNT and context parts.

\begin{figure}[t]
    \centering
    \includegraphics[width=1\linewidth]{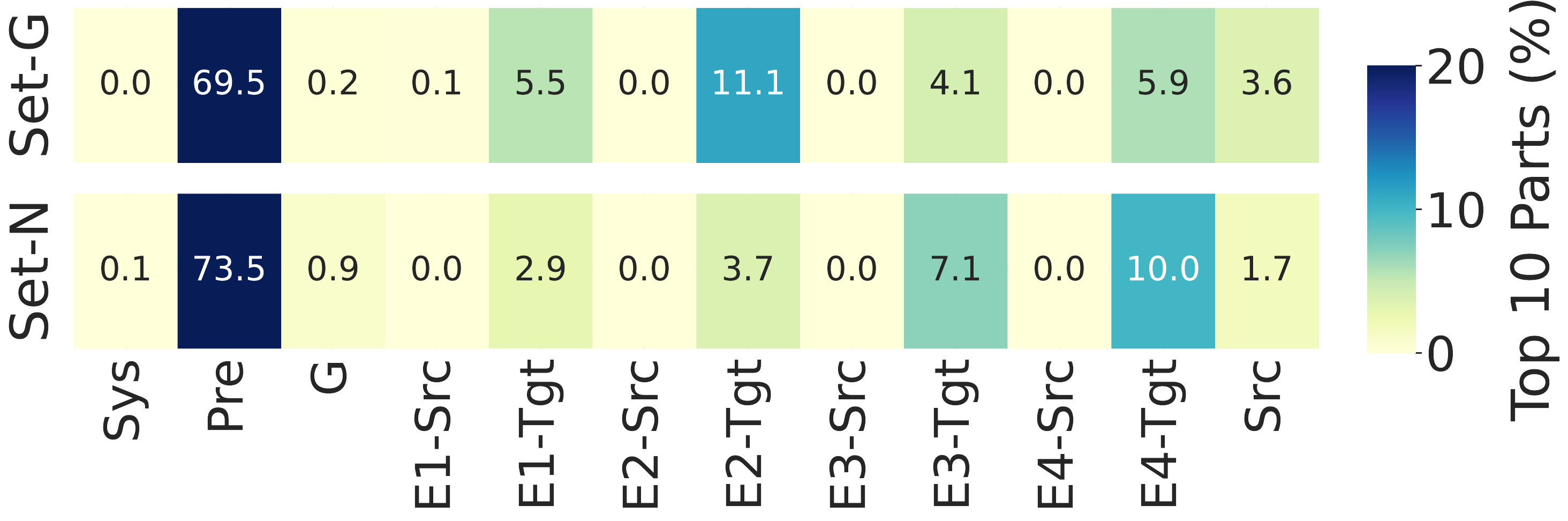}
    \caption{\textbf{Relevance of context parts to: Source Label}. Ratio of occurrence within the top 10 scores. Results for gendered (top) and neutral (bottom) source sentences.}
    \label{fig:top10-parts-label}
\end{figure}

\subsection{\bs{Attribution Setup}}
We use Attention-Aware Layer-Wise Relevance Propagation \citep[][AttnLRP]{pmlr-v235-achtibat24a}, a leading method reporting
strong faithfulness and plausibility.
Given an
input and model output, AttnLRP attributes a numerical score to each input token defined as $\text{s}(i, a:b) \in \mathbb{R}$ where $i$ is the positional index of the input token and $a$, $b$ are the (inclusive) boundaries of a span in the output (full derivation in Appendix~\ref{app:sec:context_attribution_details}). This quantity indicates the \textit{contribution of a given token for generating all the tokens within} $[a,b]$.

We compute two sets of token contributions to explain
different parts of the output:
1) \textbf{$\text{S}_{\text{L}}$} collects the contributions for generating the 
source category label 
2) \textbf{$\text{S}_{\text{T}}$} collects the scores explaining the actual translation%
---including the tokens of the label itself as 
it is
always generated before the translation. 
We exclude 
all chat template special tokens (e.g., \texttt{<|im\_start|>}) and the source tokens from $\text{S}_{\text{T}}$---expectedly, their contribution is always the strongest, preventing a meaningful interpretation of the other context parts.
For each part (the system prompt \chipbox{Sys}, preamble \chipbox{Pre}, guidelines \chipbox{G}, four task exemplars \chipbox{E}, the source sentence being translated (for $\text{S}_{\text{L}}$) \chipbox{Src}, and 
source label \chipbox{Label} when explaining $\text{S}_{\text{T}}$) we collect and aggregate token-level scores (see Appendix~\ref{app:sec:context_attribution_details} for full details and  Appendix \ref{app:xai-results} for complementary results).

\paragraph{Data} \bs{We focus on \model{qwen72} with a full prompt (G+S), the configuration that yielded the best GNT results (\S\ref{sec:gnt_results}).\footnote{We focus on one model for resource availability; the strongest one to ensure a sufficient number of GNT outputs.} 
We compute contributions on 4,000 outputs across four language pairs and both \textsc{MGeNTE} sets.  
To prioritize soundness and avoid potential errors from the automatic LLM-as-judge evaluation, \textbf{$\text{S}_{\text{T}}$} contributions are computed on output translations that are manually evaluated as gendered or neutral. All data and annotation details are provided in Appendix~\ref{app:sec:context_attribution_details}.}
\bs{Given that the number of correct label predictions and GNT outputs differs across languages (see Figure~\ref{fig:GNT-scores}), overall results in the following section are averaged across languages for equal representation.\footnote{For a breakdown by language, see Figure~\ref{fig:top_label_lang} and \ref{fig:top_translation_lang} in Appendix \ref{app:xai-results}.}}







\begin{figure}[t!]
    \centering
    \includegraphics[width=1\linewidth]{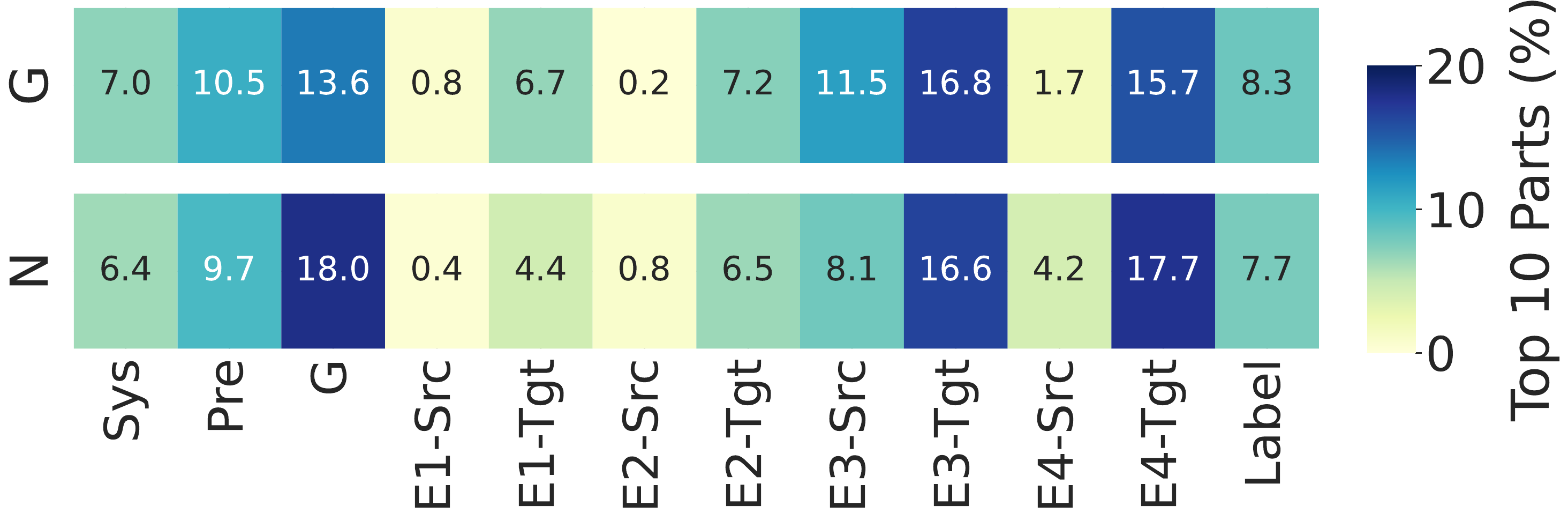}
    %
    \caption{\textbf{Relevance of context parts to: Translation}. Ratio of occurrence within the top 10 scores. Results for ambiguous sentences with  wrongly gendered (G) and correctly neutral (N) translations.}
    \label{fig:top10-trans-label}
\end{figure}

\subsection{Findings}

\paragraph{Does \model{qwen72} use the context in a similar way to \textit{detect} and \textit{translate} gender neutrality?} 
No, it does not. 
To detect the source category,
it mostly relies on the preamble \chipbox{Pre} (Figure~\ref{fig:top10-parts-label}). In contrast, contributions in $\text{S}_{\text{T}}$ are more heterogeneous (Figure~\ref{fig:top10-trans-label}), with major contributions from the guidelines (\chipbox{G}) and the assistant's neutral exemplars (\chipbox{E3/4-Tgt}).
 Asymmetry in context use can be explained by the different nature of the task, i.e., binary classification vs. open-ended generation. 

\paragraph{While \textit{detecting} the source category, which context part contributes the most relevant tokens?}
The preamble (task definition) \chipbox{Pre}, and by a large margin (Figure~\ref{fig:top10-parts-label}). The Figure shows how often the preamble is among the top 10 relevant tokens.
Given that the preamble is present across all prompt configurations, this explains the consistently high performance on the source category task. 
When observing the other top contributors, 
we see that gendered exemplars (\chipbox{E1}, \chipbox{E2}) weigh more in \textsc{Set-G}, and neutral ones in \textsc{Set-N} (\chipbox{E3}, \chipbox{E4}). 
A closer look at the label generations for \textsc{Set-N}  reveals a somewhat surprising result. Despite being accurate, \model{qwen72} does not use the input \chipbox{Src} nor the source of the shot examples (\chipbox{E3/4-Src}) when predicting the label. \chipbox{Src} contribution is marginally higher in \textsc{Set-G}, where explicit gender cues such as \textit{he}, \textit{she}, or \textit{her} appear frequently among top contributors. 
This finding suggests a lexical overfitting phenomenon: the model detects a gendered category when source gender cues are present, and assumes neutrality when they are absent.\footnote{By design, \textsc{Set-N} has no explicit source gender cues since neutrality is defined precisely by their absence.}


\paragraph{What drives \model{qwen72}'s neutral translations of ambiguous source sentences?}
The assistant's part in neutral exemplars \chipbox{E3/4-Tgt}, gender neutral guidelines \chipbox{G}, and preamble \chipbox{Pre} (Figure~\ref{fig:top10-trans-label}). 
The Figure shows how often these are among the top 10 relevant tokens, comparing wrongly gendered and correctly neutral translations from \textsc{Set-N}.
Crucially, the source label (\chipbox{Label}---always \textit{Neutral})
has a similar contribution regardless of whether the output continued with a gendered or neutral translation. 
%
This finding suggests that, in our setup, the previously generated label \textit{Neutral} (\chipbox{Label}) is not discriminative of the produced translation, whether gendered or neutral. 
This opens up to  future inquiries on 
conditioning mechanisms to link label predictions with translation behavior towards improving GNT.

\section{Conclusion}

We presented \textsc{mGeNTE}, the only existing multilingual benchmark for gender-neutral translation, covering English$\rightarrow$Italian, Spanish, German, and Greek. This expert-curated resource provides rich annotations, contrastive references, and diverse neutralization strategies to support inclusive MT research.
Using \textsc{mGeNTE}, we carried out the first systematic multilingual evaluation of open LMs for gender-neutral translation. Our findings reveal language gaps, the influence of model size, and prompt context, with interpretability analyses explaining the capabilities of LMs for this task as well as  opportunities for improvement.
Moving forward, we aim to leverage \textsc{mGeNTE}’s richness to advance fairer, more inclusive NLP \bs{and we} release it publicly to support the broader research community.




\section*{Limitations}







While \textsc{mGeNTE} represents a significant step forward in evaluating gender-neutral machine translation, we acknowledge four limitations that can be addressed in future work.

First, the findings presented in this paper are based on state-of-the-art models using both a base prompt (i.e., configuration \textit{None}, with only task definition and shots) that is incrementally made richer in information based on established prompting approaches (e.g. adding a \textit{system prompt}).  
However, it is likely that our results will not generalize to \textit{every} possible prompt formulation or model configuration for translation systems. Alternative model architectures or novel prompts may yield different performance patterns that are not captured in our current framework.

Second, \textsc{mGeNTE} employs a sentence-level design that, while effective and still quite widespread in the field, represents a simplification of real-world translation scenarios. This approach does not account for potential extra-sentential or long-range dependencies (e.g., discourse-level phenomena such as coreference resolution across multiple sentences) that may influence gender expression in translation.

Third, our best, LM-based GNT evaluation metric relies on a closed, commercial language model. While this model represents the current state-of-the-art, its proprietary nature may hinder reproducibility. Moreover, as the underlying model receives updates over time or is discontinued, our metric itself may evolve, potentially complicating longitudinal comparisons. 
Open-weight alternatives offer a promising, more transparent avenue for future research. We provide one of such in Appendix \ref{app:llm-judge}. Namely, our second-best \textit{LM evaluator} setup utilizes \model{qwen72} as a judge, achieving competitive results (approximately 7 points behind GPT-4o).  

Fourth, our interpretability analysis is conducted at a relatively coarse level, e.g., by computing attributions over the entire output translation. However, gender-translation phenomena typically manifest and regard only a portion of the overall translation output. As a result, computing attributions over the entire translation may dilute the precision of the interpretability analysis in relation to gender-specific effects. The reason for this approach, however, is that gender-neutralization often involves multiple, non-consecutive words within a sentence, making it conceptually challenging to define and isolate the relevant spans for targeted analysis. We intend to address this limitation in future work by leveraging the contrastive references provided by \textsc{mGeNTE} to enable more fine-grained, span-level attribution methods.



\section*{Ethics Statement}
\label{sec:ethic}
This paper concerns gender-neutral language and translation, inherently addressing ethical considerations. Specifically, it is intended to tackle language technologies that can generate service disparities and perpetuate exclusionary language \citep{savoldi-etal-2024-harm, ungless-etal-2025-amplifying}, thereby reinforcing stereotypes, promoting masculine dominance, and neglecting the representation of non-binary gender identities.\footnote{The term "non-binary" is used here as an inclusive umbrella term to represent identities that exist both within and beyond the masculine/feminine binary, and are not conveyed through binary linguistic expressions.}
Our focus is on gender-neutralization strategies that modify existing forms and grammatical structures to eliminate unnecessary gendered language. 
These methods aim to avoid assumptions about gender and \textit{equally represent} all gender identities in language \citep{strengers-2020}. By contrast, Direct Non-binary Language \citep{artemis}  seeks to \textit{increase} the visibility of non-binary individuals by introducing new linguistic elements such as neologisms, neopronouns, or even neomorphemes \citep{lauscher-etal-2022-welcome, rivas-ginel-2022-all-inclusive, piergentili-etal-2024-enhancing}. 

A range of strategies can be employed to meet the demand for inclusive language \citep{scharron2020latinx, Comandini_2021, knisely2020franccais}. It is crucial to note that the neutralization methods used in this research are not intended to be prescriptive. Instead, they complement other approaches and expressions for achieving inclusivity in language technologies.

\section*{Acknowledgements}

We thank Beatriz Riva, Begoña Martínez, Jens Wargenau, Pedro Cano Navarro, Sylke Denfeld, Zografia Kefalopoulou, Eleni Vlachou, and Matthew Louis Papapoulios for their essential translation work in creating the neutral references of the \textsc{mGeNTE} corpus. The translation work for \textsc{mGeNTE} en-el was funded by the Department of Translation, Interpreting and Communication of Ghent University. We also thank Verdiana Crociani for their invaluable help in the annotation and IAA computation for \textsc{mGeNTE} en-de, as well as Maria Tsaousidou and Stella Tatalia for their contribution to the annotation of \textsc{mGeNTE} en-el. 
Beatrice Savoldi is supported by the PNRR project FAIR -  Future AI Research (PE00000013),  under the NRRP MUR program funded by the NextGenerationEU. 
Matteo Negri and Luisa Bentivogli are supported by the Horizon Europe research and innovation programme,  under grant agreement No 101135798, project Meetween (My Personal AI Mediator for Virtual MEETtings BetWEEN People). 
Eleonora Cupin and Manjinder Thind's carried out their work during internships at FBK in the spring and summer of 2024, and Eleni Gkovedarou during internship at LT\textsuperscript{3} - Ghent University in the summer and winter of 2024/5. Anne Lauscher's work is funded under the excellence strategy of the German federal government and the states.
Giuseppe Attanasio is supported by the Portuguese Recovery and Resilience Plan through projects C645008882-00000055 (Center for Responsible AI) and UID/50008: Instituto de Telecomunicações.

\bibliography{anthology,custom}

\begin{thebibliography}{81}
\expandafter\ifx\csname natexlab\endcsname\relax\def\natexlab#1{#1}\fi

\bibitem[{Abdin et~al.(2024)Abdin, Aneja, Behl, Bubeck, Eldan, Gunasekar, Harrison, Hewett, Javaheripi, Kauffmann et~al.}]{abdin2024phi4technicalreport}
Marah Abdin, Jyoti Aneja, Harkirat Behl, Sébastien Bubeck, Ronen Eldan, Suriya Gunasekar, Michael Harrison, Russell~J. Hewett, Mojan Javaheripi, Piero Kauffmann, et~al. 2024.
\newblock \href {http://arxiv.org/abs/2412.08905} {Phi-4 technical report}.

\bibitem[{Achtibat et~al.(2024)Achtibat, Hatefi, Dreyer, Jain, Wiegand, Lapuschkin, and Samek}]{pmlr-v235-achtibat24a}
Reduan Achtibat, Sayed Mohammad~Vakilzadeh Hatefi, Maximilian Dreyer, Aakriti Jain, Thomas Wiegand, Sebastian Lapuschkin, and Wojciech Samek. 2024.
\newblock {A}ttn{LRP}: Attention-aware layer-wise relevance propagation for transformers.
\newblock In \emph{Proceedings of the 41st International Conference on Machine Learning}, volume 235 of \emph{Proceedings of Machine Learning Research}, pages 135--168. PMLR.

\bibitem[{Amrhein et~al.(2023)Amrhein, Schottmann, Sennrich, and L{\"a}ubli}]{amrhein-etal-2023-exploiting}
Chantal Amrhein, Florian Schottmann, Rico Sennrich, and Samuel L{\"a}ubli. 2023.
\newblock \href {https://doi.org/10.18653/v1/2023.acl-long.246} {Exploiting biased models to de-bias text: A gender-fair rewriting model}.
\newblock In \emph{Proceedings of the 61st Annual Meeting of the Association for Computational Linguistics (Volume 1: Long Papers)}, pages 4486--4506, Toronto, Canada. Association for Computational Linguistics.

\bibitem[{APA(2020)}]{apa2020publication}
APA. 2020.
\newblock \emph{Publication Manual of the American Psychological Association}, 7th edition.
\newblock American Psychological Association.

\bibitem[{Arras et~al.(2025)Arras, Puri, Kahardipraja, Lapuschkin, and Samek}]{arras2025close}
Leila Arras, Bruno Puri, Patrick Kahardipraja, Sebastian Lapuschkin, and Wojciech Samek. 2025.
\newblock A close look at decomposition-based xai-methods for transformer language models.
\newblock \emph{arXiv preprint arXiv:2502.15886}.

\bibitem[{Ashwell et~al.(2023)Ashwell, Baskin, Christiansen, DiBari, and Flanagin}]{Ashwell2023}
Sabrina~J. Ashwell, Patricia~K. Baskin, Stacy~L. Christiansen, Sara~A. DiBari, and Annette Flanagin. 2023.
\newblock \href {https://doi.org/10.1002/leap.1527} {Three recommended inclusive language guidelines for scholarly publishing: Words matter}.
\newblock \emph{Learned Publishing}, 36(1):94--99.

\bibitem[{Attanasio et~al.(2021)Attanasio, Greco, La~Quatra, Cagliero, Tonti, Cerquitelli, and Raus}]{attanasio2021mimic}
Giuseppe Attanasio, Salvatore Greco, Moreno La~Quatra, Luca Cagliero, Michela Tonti, Tania Cerquitelli, and Rachele Raus. 2021.
\newblock E-mimic: Empowering multilingual inclusive communication.
\newblock In \emph{2021 IEEE International Conference on Big Data (Big Data)}, pages 4227--4234. IEEE.

\bibitem[{Attanasio et~al.(2023)Attanasio, Plaza~del Arco, Nozza, and Lauscher}]{attanasio-etal-2023-tale}
Giuseppe Attanasio, Flor~Miriam Plaza~del Arco, Debora Nozza, and Anne Lauscher. 2023.
\newblock \href {https://doi.org/10.18653/v1/2023.emnlp-main.243} {A tale of pronouns: Interpretability informs gender bias mitigation for fairer instruction-tuned machine translation}.
\newblock In \emph{Proceedings of the 2023 Conference on Empirical Methods in Natural Language Processing}, pages 3996--4014, Singapore. Association for Computational Linguistics.

\bibitem[{Bartl and Leavy(2024)}]{bartl-leavy-2024-showgirls}
Marion Bartl and Susan Leavy. 2024.
\newblock \href {https://doi.org/10.18653/v1/2024.gebnlp-1.18} {From {`}showgirls{'} to {`}performers{'}: Fine-tuning with gender-inclusive language for bias reduction in {LLM}s}.
\newblock In \emph{Proceedings of the 5th Workshop on Gender Bias in Natural Language Processing (GeBNLP)}, pages 280--294, Bangkok, Thailand. Association for Computational Linguistics.

\bibitem[{Bartl et~al.(2025)Bartl, Murphy, and Leavy}]{bartl2025adapting}
Marion Bartl, Thomas~Brendan Murphy, and Susan Leavy. 2025.
\newblock Adapting psycholinguistic research for llms: Gender-inclusive language in a coreference context.
\newblock \emph{arXiv preprint arXiv:2502.13120}.

\bibitem[{Brown et~al.(2020)Brown, Mann, Ryder, Subbiah, Kaplan, Dhariwal, Neelakantan, Shyam, Sastry, Askell, Agarwal, Herbert-Voss, Krueger, Henighan, Child, Ramesh, Ziegler, Wu, Winter, Hesse, Chen, Sigler, Litwin, Gray, Chess, Clark, Berner, McCandlish, Radford, Sutskever, and Amodei}]{brown}
Tom~B. Brown, Benjamin Mann, Nick Ryder, Melanie Subbiah, Jared Kaplan, Prafulla Dhariwal, Arvind Neelakantan, Pranav Shyam, Girish Sastry, Amanda Askell, Sandhini Agarwal, Ariel Herbert-Voss, Gretchen Krueger, Tom Henighan, Rewon Child, Aditya Ramesh, Daniel~M. Ziegler, Jeffrey Wu, Clemens Winter, Christopher Hesse, Mark Chen, Eric Sigler, Mateusz Litwin, Scott Gray, Benjamin Chess, Jack Clark, Christopher Berner, Sam McCandlish, Alec Radford, Ilya Sutskever, and Dario Amodei. 2020.
\newblock Language models are few-shot learners.
\newblock In \emph{Proceedings of the 34th International Conference on Neural Information Processing Systems}, NIPS'20, Red Hook, NY, USA. Curran Associates Inc.

\bibitem[{Bunzeck and Zarrie{\ss}(2024)}]{bunzeck-zarriess-2024-slayqa}
Bastian Bunzeck and Sina Zarrie{\ss}. 2024.
\newblock \href {https://doi.org/10.18653/v1/2024.genbench-1.3} {The {S}lay{QA} benchmark of social reasoning: testing gender-inclusive generalization with neopronouns}.
\newblock In \emph{Proceedings of the 2nd GenBench Workshop on Generalisation (Benchmarking) in NLP}, pages 42--53, Miami, Florida, USA. Association for Computational Linguistics.

\bibitem[{Cao and Daum{\'e}~III(2020)}]{cao-daume-iii-2020-toward}
Yang~Trista Cao and Hal Daum{\'e}~III. 2020.
\newblock \href {https://doi.org/10.18653/v1/2020.acl-main.418} {Toward gender-inclusive coreference resolution}.
\newblock In \emph{Proceedings of the 58th Annual Meeting of the Association for Computational Linguistics}, pages 4568--4595, Online. Association for Computational Linguistics.

\bibitem[{Cho et~al.(2019)Cho, Kim, Kim, and Kim}]{cho-etal-2019measuring}
Won~Ik Cho, Ji~Won Kim, Seok~Min Kim, and Nam~Soo Kim. 2019.
\newblock \href {https://doi.org/10.18653/v1/W19-3824} {On {M}easuring {G}ender bias in {T}ranslation of {G}ender-neutral {P}ronouns}.
\newblock In \emph{Proceedings of the First Workshop on Gender Bias in Natural Language Processing}, pages 173--181, Florence, IT. Association for Computational Linguistics.

\bibitem[{Cohen-Wang et~al.(2024)Cohen-Wang, Shah, Georgiev, and Madry}]{cohen2024contextcite}
Benjamin Cohen-Wang, Harshay Shah, Kristian Georgiev, and Aleksander Madry. 2024.
\newblock Contextcite: Attributing model generation to context.
\newblock \emph{Advances in Neural Information Processing Systems}, 37:95764--95807.

\bibitem[{Comandini(2021)}]{Comandini_2021}
Gloria Comandini. 2021.
\newblock \href {https://testoesenso.it/index.php/testoesenso/article/view/524} {Salve a tutt\textrev{e}, tutt*, tuttu, tuttx e tutt@: l’uso delle strategie di neutralizzazione di genere nella comunità queer online}.
\newblock \emph{Testo e Senso}, 23:43–64.

\bibitem[{Daems(2023)}]{daems-2023-gender}
Joke Daems. 2023.
\newblock \href {https://aclanthology.org/2023.gitt-1.4} {Gender-inclusive translation for a gender-inclusive sport: strategies and translator perceptions at the international quadball association}.
\newblock In \emph{Proceedings of the First Workshop on Gender-Inclusive Translation Technologies}, pages 37--47, Tampere, Finland. European Association for Machine Translation.

\bibitem[{di~Carlo(2024)}]{ScottoDiCarlo2024}
Giuseppina~Scotto di~Carlo. 2024.
\newblock \href {https://doi.org/10.4324/9781003432906} {Is italy ready for gender-inclusive language? an attitude and usage study among italian speakers}.
\newblock In \emph{Inclusiveness Beyond the (Non)binary in Romance Languages}, 1st edition edition, page~21. Routledge.

\bibitem[{Dice(1945)}]{dice1945measures}
Lee~R. Dice. 1945.
\newblock {Measures of the Amount of Ecologic Association Between Species}.
\newblock \emph{Ecology}, 26(3):297--302.

\bibitem[{Ferrando et~al.(2022)Ferrando, G{\'a}llego, Alastruey, Escolano, and Costa-juss{\`a}}]{ferrando-etal-2022-towards}
Javier Ferrando, Gerard~I. G{\'a}llego, Belen Alastruey, Carlos Escolano, and Marta~R. Costa-juss{\`a}. 2022.
\newblock \href {https://doi.org/10.18653/v1/2022.emnlp-main.599} {Towards opening the black box of neural machine translation: Source and target interpretations of the transformer}.
\newblock In \emph{Proceedings of the 2022 Conference on Empirical Methods in Natural Language Processing}, pages 8756--8769, Abu Dhabi, United Arab Emirates. Association for Computational Linguistics.

\bibitem[{Frenda et~al.(2024)Frenda, Piergentili, Savoldi, Madeddu, Rosola, Casola, Ferrando, Patti, Negri, Bentivogli et~al.}]{frenda2024gfg}
Simona Frenda, Andrea Piergentili, Beatrice Savoldi, Marco Madeddu, Martina Rosola, Silvia Casola, Chiara Ferrando, Viviana Patti, Matteo Negri, Luisa Bentivogli, et~al. 2024.
\newblock Gfg-gender-fair generation: A calamita challenge.
\newblock In \emph{Proceedings of the Tenth Italian Conference on Computational Linguistics (CLiC-it 2024)}.

\bibitem[{Friidhriksd{\'o}ttir(2024)}]{friidhriksdottir-2024-genderqueer}
Steinunn~Rut Friidhriksd{\'o}ttir. 2024.
\newblock \href {https://doi.org/10.18653/v1/2024.wmt-1.26} {The {G}ender{Q}ueer test suite}.
\newblock In \emph{Proceedings of the Ninth Conference on Machine Translation}, pages 327--340, Miami, Florida, USA. Association for Computational Linguistics.

\bibitem[{Gaido et~al.(2020)Gaido, Savoldi, Bentivogli, Negri, and Turchi}]{gaido-etal-2020-breeding}
Marco Gaido, Beatrice Savoldi, Luisa Bentivogli, Matteo Negri, and Marco Turchi. 2020.
\newblock \href {https://doi.org/10.18653/v1/2020.coling-main.350} {Breeding gender-aware direct speech translation systems}.
\newblock In \emph{Proceedings of the 28th International Conference on Computational Linguistics}, pages 3951--3964, Barcelona, Spain (Online). International Committee on Computational Linguistics.

\bibitem[{Gautam et~al.(2024)Gautam, Bingert, Zhu, Lauscher, and Klakow}]{gautam2024}
Vagrant Gautam, Eileen Bingert, Dawei Zhu, Anne Lauscher, and Dietrich Klakow. 2024.
\newblock \href {https://doi.org/10.1162/tacl_a_00719} {Robust pronoun fidelity with english llms: Are they reasoning, repeating, or just biased?}
\newblock \emph{Transactions of the Association for Computational Linguistics}, 12:1755--1779.

\bibitem[{Ginel and Theroine(2022)}]{rivas-ginel-2022-all-inclusive}
María Isabel~Rivas Ginel and Sarah Theroine. 2022.
\newblock Neutralising for equality: All-inclusive games machine translation.
\newblock In \emph{Proceedings of New Trends in Translation and Technology}, pages 125--133. NeTTT.

\bibitem[{Grattafiori et~al.(2024)Grattafiori, Dubey, Jauhri, Pandey, Kadian, Al-Dahle, Letman, Mathur, Schelten, Vaughan et~al.}]{grattafiori2024llama3herdmodels}
Aaron Grattafiori, Abhimanyu Dubey, Abhinav Jauhri, Abhinav Pandey, Abhishek Kadian, Ahmad Al-Dahle, Aiesha Letman, Akhil Mathur, Alan Schelten, Alex Vaughan, et~al. 2024.
\newblock \href {http://arxiv.org/abs/2407.21783} {The llama 3 herd of models}.

\bibitem[{Greco et~al.(2025)Greco, La~Quatra, Cagliero, and Cerquitelli}]{greco-emimic}
Salvatore Greco, Moreno La~Quatra, Luca Cagliero, and Tania Cerquitelli. 2025.
\newblock \href {https://doi.org/10.1145/3729237} {Towards ai-assisted inclusive language writing in italian formal communications}.
\newblock \emph{ACM Trans. Intell. Syst. Technol.}

\bibitem[{Gu et~al.(2025)Gu, Jiang, Shi, Tan, Zhai, Xu, Li, Shen, Ma, Liu, Wang, Zhang, Wang, Gao, Ni, and Guo}]{gu2025surveyllmasajudge}
Jiawei Gu, Xuhui Jiang, Zhichao Shi, Hexiang Tan, Xuehao Zhai, Chengjin Xu, Wei Li, Yinghan Shen, Shengjie Ma, Honghao Liu, Saizhuo Wang, Kun Zhang, Yuanzhuo Wang, Wen Gao, Lionel Ni, and Jian Guo. 2025.
\newblock \href {http://arxiv.org/abs/2411.15594} {A survey on llm-as-a-judge}.

\bibitem[{Guerreiro et~al.(2024)Guerreiro, Rei, Stigt, Coheur, Colombo, and Martins}]{guerreiro-etal-2024-xcomet}
Nuno~M. Guerreiro, Ricardo Rei, Daan~van Stigt, Luisa Coheur, Pierre Colombo, and Andr{\'e} F.~T. Martins. 2024.
\newblock \href {https://doi.org/10.1162/tacl_a_00683} {xcomet: Transparent machine translation evaluation through fine-grained error detection}.
\newblock \emph{Transactions of the Association for Computational Linguistics}, 12:979--995.

\bibitem[{Gygax et~al.(2021)Gygax, Sato, {\"O}ttl, and Gabriel}]{gygax2021masculine}
Pascal Gygax, Sayaka Sato, Anton {\"O}ttl, and Ute Gabriel. 2021.
\newblock The masculine form in grammatically gendered languages and its multiple interpretations: A challenge for our cognitive system.
\newblock \emph{Language Sciences}, 83:101328.

\bibitem[{Gygax et~al.(2019)Gygax, Elmiger, Zufferey, Garnham, Sczesny, von Stockhausen, Braun, and Oakhill}]{Gygaxindex4}
Pascal~M. Gygax, Daniel Elmiger, Sandrine Zufferey, Alan Garnham, Sabine Sczesny, Lisa von Stockhausen, Friederike Braun, and Jane Oakhill. 2019.
\newblock \href {https://doi.org/10.3389/fpsyg.2019.01604} {{A Language Index of Grammatical Gender Dimensions to Study the Impact of Grammatical Gender on the Way We Perceive Women and Men}}.
\newblock \emph{Frontiers in Psychology}, 10:1604.

\bibitem[{H{\"o}glund and Flinkfeldt(2023)}]{hoglund2023gendering}
Frida H{\"o}glund and Marie Flinkfeldt. 2023.
\newblock De-gendering parents: Gender inclusion and standardised language in screen-level bureaucracy.
\newblock \emph{International Journal of Social Welfare}.

\bibitem[{Hossain et~al.(2023)Hossain, Dev, and Singh}]{hossain-etal-2023-misgendered}
Tamanna Hossain, Sunipa Dev, and Sameer Singh. 2023.
\newblock \href {https://doi.org/10.18653/v1/2023.acl-long.293} {{MISGENDERED}: Limits of large language models in understanding pronouns}.
\newblock In \emph{Proceedings of the 61st Annual Meeting of the Association for Computational Linguistics (Volume 1: Long Papers)}, pages 5352--5367, Toronto, Canada. Association for Computational Linguistics.

\bibitem[{Jourdan et~al.(2025)Jourdan, Chevalier, and Favre}]{jourdan2025fairtranslate}
Fanny Jourdan, Yannick Chevalier, and C{\'e}cile Favre. 2025.
\newblock Fairtranslate: An english-french dataset for gender bias evaluation in machine translation by overcoming gender binarity.
\newblock \emph{arXiv preprint arXiv:2504.15941}.

\bibitem[{Knisely(2020)}]{knisely2020franccais}
Kris~Aric Knisely. 2020.
\newblock \href {https://doi.org/https://doi.org/10.1111/flan.12500} {{Le fran{\c{c}}ais non-binaire: Linguistic forms used by non-binary speakers of French}}.
\newblock \emph{Foreign Language Annals}, 53(4):850--876.

\bibitem[{Koehn(2005)}]{koehn2005europarl}
Philipp Koehn. 2005.
\newblock \href {http://mt-archive.info/MTS-2005-Koehn.pdf} {{Europarl: A Parallel Corpus for Statistical Machine Translation}}.
\newblock In \emph{{Proceedings of the tenth Machine Translation Summit}}, pages 79--86, Phuket, TH. AAMT.

\bibitem[{Kwon et~al.(2023)Kwon, Li, Zhuang, Sheng, Zheng, Yu, Gonzalez, Zhang, and Stoica}]{kwon2023efficient}
Woosuk Kwon, Zhuohan Li, Siyuan Zhuang, Ying Sheng, Lianmin Zheng, Cody~Hao Yu, Joseph~E. Gonzalez, Hao Zhang, and Ion Stoica. 2023.
\newblock Efficient memory management for large language model serving with pagedattention.
\newblock In \emph{Proceedings of the ACM SIGOPS 29th Symposium on Operating Systems Principles}.

\bibitem[{Lardelli et~al.(2024{\natexlab{a}})Lardelli, Attanasio, and Lauscher}]{lardelli-etal-2024-building}
Manuel Lardelli, Giuseppe Attanasio, and Anne Lauscher. 2024{\natexlab{a}}.
\newblock \href {https://doi.org/10.18653/v1/2024.findings-acl.448} {Building bridges: A dataset for evaluating gender-fair machine translation into {G}erman}.
\newblock In \emph{Findings of the Association for Computational Linguistics: ACL 2024}, pages 7542--7550, Bangkok, Thailand. Association for Computational Linguistics.

\bibitem[{Lardelli et~al.(2024{\natexlab{b}})Lardelli, Dill, Attanasio, and Lauscher}]{lardelli-etal-2024-sparks}
Manuel Lardelli, Timm Dill, Giuseppe Attanasio, and Anne Lauscher. 2024{\natexlab{b}}.
\newblock \href {https://aclanthology.org/2024.gitt-1.2} {Sparks of fairness: Preliminary evidence of commercial machine translation as {E}nglish-to-{G}erman gender-fair dictionaries}.
\newblock In \emph{Proceedings of the 2nd International Workshop on Gender-Inclusive Translation Technologies}, pages 12--21, Sheffield, United Kingdom. European Association for Machine Translation (EAMT).

\bibitem[{Lauscher et~al.(2022)Lauscher, Crowley, and Hovy}]{lauscher-etal-2022-welcome}
Anne Lauscher, Archie Crowley, and Dirk Hovy. 2022.
\newblock \href {https://aclanthology.org/2022.coling-1.105} {Welcome to the modern world of pronouns: Identity-inclusive natural language processing beyond gender}.
\newblock In \emph{Proceedings of the 29th International Conference on Computational Linguistics}, pages 1221--1232, Gyeongju, Republic of Korea. International Committee on Computational Linguistics.

\bibitem[{Lauscher et~al.(2023)Lauscher, Nozza, Miltersen, Crowley, and Hovy}]{lauscher-etal-2023-em}
Anne Lauscher, Debora Nozza, Ehm Miltersen, Archie Crowley, and Dirk Hovy. 2023.
\newblock \href {https://doi.org/10.18653/v1/2023.acl-long.23} {What about {``}em{''}? how commercial machine translation fails to handle (neo-)pronouns}.
\newblock In \emph{Proceedings of the 61st Annual Meeting of the Association for Computational Linguistics (Volume 1: Long Papers)}, pages 377--392, Toronto, Canada. Association for Computational Linguistics.

\bibitem[{Liu et~al.(2024)Liu, Kandpal, and Raffel}]{liu2024attribot}
Fengyuan Liu, Nikhil Kandpal, and Colin Raffel. 2024.
\newblock Attribot: A bag of tricks for efficiently approximating leave-one-out context attribution.
\newblock \emph{arXiv preprint arXiv:2411.15102}.

\bibitem[{López(2020)}]{artemis}
Ártemis López. 2020.
\newblock \href {https://fh.mdp.edu.ar/revistas/index.php/cuarentanaipes/article/view/4891} {Cuando el lenguaje excluye: Consideraciones sobre el lenguaje no binario indirecto}.
\newblock \emph{Cuarenta Naipes}, 3:295--312.

\bibitem[{Madsen et~al.(2022)Madsen, Reddy, and Chandar}]{madsen2022post}
Andreas Madsen, Siva Reddy, and Sarath Chandar. 2022.
\newblock Post-hoc interpretability for neural nlp: A survey.
\newblock \emph{ACM Computing Surveys}, 55(8):1--42.

\bibitem[{Mosca et~al.(2022)Mosca, Szigeti, Tragianni, Gallagher, and Groh}]{mosca-etal-2022-shap}
Edoardo Mosca, Ferenc Szigeti, Stella Tragianni, Daniel Gallagher, and Georg Groh. 2022.
\newblock \href {https://aclanthology.org/2022.coling-1.406} {{SHAP}-based explanation methods: A review for {NLP} interpretability}.
\newblock In \emph{Proceedings of the 29th International Conference on Computational Linguistics}, pages 4593--4603, Gyeongju, Republic of Korea. International Committee on Computational Linguistics.

\bibitem[{M{\"u}ller-Spitzer et~al.(2024)M{\"u}ller-Spitzer, Ochs, Koplenig, R{\"u}diger, and Wolfer}]{muller2024less}
Carolin M{\"u}ller-Spitzer, Samira Ochs, Alexander Koplenig, Jan~Oliver R{\"u}diger, and Sascha Wolfer. 2024.
\newblock Less than one percent of words would be affected by gender-inclusive language in german press texts.
\newblock \emph{Humanities and Social Sciences Communications}, 11(1):1--13.

\bibitem[{Ovalle et~al.(2023)Ovalle, Goyal, Dhamala, Jaggers, Chang, Galstyan, Zemel, and Gupta}]{ovalle2023-towards-centering}
Anaelia Ovalle, Palash Goyal, Jwala Dhamala, Zachary Jaggers, Kai-Wei Chang, Aram Galstyan, Richard Zemel, and Rahul Gupta. 2023.
\newblock \href {https://doi.org/10.1145/3593013.3594078} {“i’m fully who i am”: Towards centering transgender and non-binary voices to measure biases in open language generation}.
\newblock In \emph{Proceedings of the 2023 ACM Conference on Fairness, Accountability, and Transparency}, FAccT '23, page 1246–1266, New York, NY, USA. Association for Computing Machinery.

\bibitem[{Papadopoulos(2022)}]{papadopoulos2022brief}
Ben Papadopoulos. 2022.
\newblock {A brief history of gender-inclusive Spanish}.
\newblock \emph{Deportate, esuli, profughe}, 48(1):31--48.

\bibitem[{Piergentili et~al.(2023{\natexlab{a}})Piergentili, Fucci, Savoldi, Bentivogli, and Negri}]{piergentili2023gender}
Andrea Piergentili, Dennis Fucci, Beatrice Savoldi, Luisa Bentivogli, and Matteo Negri. 2023{\natexlab{a}}.
\newblock \href {http://arxiv.org/abs/2301.10075} {Gender neutralization for an inclusive machine translation: from theoretical foundations to open challenges}.

\bibitem[{Piergentili et~al.(2023{\natexlab{b}})Piergentili, Savoldi, Fucci, Negri, and Bentivogli}]{piergentili-etal-2023-hi}
Andrea Piergentili, Beatrice Savoldi, Dennis Fucci, Matteo Negri, and Luisa Bentivogli. 2023{\natexlab{b}}.
\newblock \href {https://doi.org/10.18653/v1/2023.emnlp-main.873} {Hi guys or hi folks? benchmarking gender-neutral machine translation with the {G}e{NTE} corpus}.
\newblock In \emph{Proceedings of the 2023 Conference on Empirical Methods in Natural Language Processing}, pages 14124--14140, Singapore. Association for Computational Linguistics.

\bibitem[{Piergentili et~al.(2024)Piergentili, Savoldi, Negri, and Bentivogli}]{piergentili-etal-2024-enhancing}
Andrea Piergentili, Beatrice Savoldi, Matteo Negri, and Luisa Bentivogli. 2024.
\newblock \href {https://aclanthology.org/2024.eamt-1.25} {Enhancing gender-inclusive machine translation with neomorphemes and large language models}.
\newblock In \emph{Proceedings of the 25th Annual Conference of the European Association for Machine Translation (Volume 1)}, pages 300--314, Sheffield, UK. European Association for Machine Translation (EAMT).

\bibitem[{Piergentili et~al.(2025)Piergentili, Savoldi, Negri, and Bentivogli}]{piergentili2025llmasajudgeapproachscalablegenderneutral}
Andrea Piergentili, Beatrice Savoldi, Matteo Negri, and Luisa Bentivogli. 2025.
\newblock \href {https://aclanthology.org/2025.gitt-1.3/} {An {LLM}-as-a-judge approach for scalable gender-neutral translation evaluation}.
\newblock In \emph{Proceedings of the 3rd Workshop on Gender-Inclusive Translation Technologies (GITT 2025)}, pages 46--63, Geneva, Switzerland. European Association for Machine Translation.

\bibitem[{Qwen(2024)}]{qwen2024}
Qwen. 2024.
\newblock Qwen2.5: A party of foundation models!
\newblock \url{https://qwenlm.github.io/blog/qwen2.5/}.
\newblock Accessed: 2025-05-03.

\bibitem[{Rei et~al.(2021)Rei, Farinha, Zerva, van Stigt, Stewart, Ramos, Glushkova, Martins, and Lavie}]{rei-etal-2021-references}
Ricardo Rei, Ana~C Farinha, Chrysoula Zerva, Daan van Stigt, Craig Stewart, Pedro Ramos, Taisiya Glushkova, Andr{\'e} F.~T. Martins, and Alon Lavie. 2021.
\newblock \href {https://aclanthology.org/2021.wmt-1.111} {Are references really needed? unbabel-{IST} 2021 submission for the metrics shared task}.
\newblock In \emph{Proceedings of the Sixth Conference on Machine Translation}, pages 1030--1040, Online. Association for Computational Linguistics.

\bibitem[{Sarti et~al.(2024)Sarti, Chrupa{\l}a, Nissim, and Bisazza}]{sarti2024quantifying}
Gabriele Sarti, Grzegorz Chrupa{\l}a, Malvina Nissim, and Arianna Bisazza. 2024.
\newblock \href {https://openreview.net/forum?id=XTHfNGI3zT} {Quantifying the plausibility of context reliance in neural machine translation}.
\newblock In \emph{The Twelfth International Conference on Learning Representations}.

\bibitem[{Sarti et~al.(2023)Sarti, Feldhus, Sickert, and van~der Wal}]{sarti-etal-2023-inseq}
Gabriele Sarti, Nils Feldhus, Ludwig Sickert, and Oskar van~der Wal. 2023.
\newblock \href {https://doi.org/10.18653/v1/2023.acl-demo.40} {Inseq: An interpretability toolkit for sequence generation models}.
\newblock In \emph{Proceedings of the 61st Annual Meeting of the Association for Computational Linguistics (Volume 3: System Demonstrations)}, pages 421--435, Toronto, Canada. Association for Computational Linguistics.

\bibitem[{Saunders(2022)}]{saunders-2022-domain}
Danielle Saunders. 2022.
\newblock \href {https://aclanthology.org/2022.eamt-1.3} {Domain adaptation for neural machine translation}.
\newblock In \emph{Proceedings of the 23rd Annual Conference of the European Association for Machine Translation}, pages 9--10, Ghent, Belgium. European Association for Machine Translation.

\bibitem[{Saunders and Byrne(2020)}]{saunders-byrne-2020-addressing}
Danielle Saunders and Bill Byrne. 2020.
\newblock \href {https://aclanthology.org/2020.wmt-1.94} {Addressing exposure bias with document minimum risk training: {C}ambridge at the {WMT}20 biomedical translation task}.
\newblock In \emph{Proceedings of the Fifth Conference on Machine Translation}, pages 862--869, Online. Association for Computational Linguistics.

\bibitem[{Saunders et~al.(2020)Saunders, Sallis, and Byrne}]{saunders-etal-2020-neural}
Danielle Saunders, Rosie Sallis, and Bill Byrne. 2020.
\newblock \href {https://aclanthology.org/2020.gebnlp-1.4} {Neural machine translation doesn{'}t translate gender coreference right unless you make it}.
\newblock In \emph{Proceedings of the Second Workshop on Gender Bias in Natural Language Processing}, pages 35--43, Barcelona, Spain (Online). Association for Computational Linguistics.

\bibitem[{Savoldi et~al.(2025)Savoldi, Bastings, Bentivogli, and Vanmassenhove}]{SAVOLDI2025101257}
Beatrice Savoldi, Jasmijn Bastings, Luisa Bentivogli, and Eva Vanmassenhove. 2025.
\newblock \href {https://doi.org/https://doi.org/10.1016/j.patter.2025.101257} {A decade of gender bias in machine translation}.
\newblock \emph{Patterns}, page 101257.

\bibitem[{Savoldi et~al.(2023)Savoldi, Gaido, Negri, and Bentivogli}]{savoldi-etal-2023-test}
Beatrice Savoldi, Marco Gaido, Matteo Negri, and Luisa Bentivogli. 2023.
\newblock \href {https://doi.org/10.18653/v1/2023.wmt-1.25} {Test suites task: Evaluation of gender fairness in {MT} with {M}u{ST}-{SHE} and {INES}}.
\newblock In \emph{Proceedings of the Eighth Conference on Machine Translation}, pages 252--262, Singapore. Association for Computational Linguistics.

\bibitem[{Savoldi et~al.(2024{\natexlab{a}})Savoldi, Papi, Negri, Guerberof-Arenas, and Bentivogli}]{savoldi-etal-2024-harm}
Beatrice Savoldi, Sara Papi, Matteo Negri, Ana Guerberof-Arenas, and Luisa Bentivogli. 2024{\natexlab{a}}.
\newblock \href {https://doi.org/10.18653/v1/2024.emnlp-main.1002} {What the harm? quantifying the tangible impact of gender bias in machine translation with a human-centered study}.
\newblock In \emph{Proceedings of the 2024 Conference on Empirical Methods in Natural Language Processing}, pages 18048--18076, Miami, Florida, USA. Association for Computational Linguistics.

\bibitem[{Savoldi et~al.(2024{\natexlab{b}})Savoldi, Piergentili, Fucci, Negri, and Bentivogli}]{savoldi-etal-2024-prompt}
Beatrice Savoldi, Andrea Piergentili, Dennis Fucci, Matteo Negri, and Luisa Bentivogli. 2024{\natexlab{b}}.
\newblock \href {https://aclanthology.org/2024.eacl-short.23} {A prompt response to the demand for automatic gender-neutral translation}.
\newblock In \emph{Proceedings of the 18th Conference of the European Chapter of the Association for Computational Linguistics (Volume 2: Short Papers)}, pages 256--267, St. Julian{'}s, Malta. Association for Computational Linguistics.

\bibitem[{Scharr{\'o}n-del R{\'\i}o and Aja(2020)}]{scharron2020latinx}
Mar{\'\i}a~R Scharr{\'o}n-del R{\'\i}o and Alan~A Aja. 2020.
\newblock Latinx: Inclusive language as liberation praxis.
\newblock \emph{Journal of Latinx Psychology}, 8(1):7.

\bibitem[{Sczesny et~al.(2016)Sczesny, Formanowicz, and Moser}]{sczesny2016can}
Sabine Sczesny, Magda Formanowicz, and Franziska Moser. 2016.
\newblock Can gender-fair language reduce gender stereotyping and discrimination?
\newblock \emph{Frontiers in psychology}, 7:154379.

\bibitem[{Shrikumar et~al.(2017)Shrikumar, Greenside, and Kundaje}]{pmlr-v70-shrikumar17a}
Avanti Shrikumar, Peyton Greenside, and Anshul Kundaje. 2017.
\newblock \href {https://proceedings.mlr.press/v70/shrikumar17a.html} {Learning important features through propagating activation differences}.
\newblock In \emph{Proceedings of the 34th International Conference on Machine Learning}, volume~70 of \emph{Proceedings of Machine Learning Research}, pages 3145--3153. PMLR.

\bibitem[{Silva and Soares(2024)}]{SilvaSoares2024}
Gláucia~V. Silva and Cristiane Soares, editors. 2024.
\newblock \href {https://doi.org/10.4324/9781003432906} {\emph{Inclusiveness Beyond the (Non)binary in Romance Languages: Research and Classroom Implementation}}, 1st edition edition.
\newblock Routledge.

\bibitem[{Silveira(1980)}]{silveira1980generic}
Jeanette Silveira. 1980.
\newblock \href {https://www.sciencedirect.com/science/article/pii/S0148068580921132} {Generic {M}asculine {W}ords and {T}hinking}.
\newblock \emph{Women's Studies International Quarterly}, 3(2-3):165--178.

\bibitem[{Stahlberg et~al.(2007)Stahlberg, Braun, Irmen, and Sczesny}]{stahlberg2007representation}
Dagmar Stahlberg, Friederike Braun, Lisa Irmen, and Sabine Sczesny. 2007.
\newblock \href {https://psycnet.apa.org/record/2007-01308-006} {{Representation of the Sexes in Language}}.
\newblock \emph{Social communication}, pages 163--187.

\bibitem[{Strengers et~al.(2020)Strengers, Qu, Xu, and Knibbe}]{strengers-2020}
Yolande Strengers, Lizhen Qu, Qiongkai Xu, and Jarrod Knibbe. 2020.
\newblock \href {https://doi.org/10.1145/3313831.3376315} {Adhering, steering, and queering: Treatment of gender in natural language generation}.
\newblock In \emph{Proceedings of the 2020 CHI Conference on Human Factors in Computing Systems}, CHI '20, page 1–14, New York, NY, USA. Association for Computing Machinery.

\bibitem[{Subramonian et~al.(2025)Subramonian, Gautam, Seshadri, Klakow, Chang, and Sun}]{subramonian2025agree}
Arjun Subramonian, Vagrant Gautam, Preethi Seshadri, Dietrich Klakow, Kai-Wei Chang, and Yizhou Sun. 2025.
\newblock Agree to disagree? a meta-evaluation of llm misgendering.
\newblock \emph{arXiv preprint arXiv:2504.17075}.

\bibitem[{Sun et~al.(2021)Sun, Webster, Shah, Wang, and Johnson}]{sun2021they}
Tony Sun, Kellie Webster, Apu Shah, William~Yang Wang, and Melvin Johnson. 2021.
\newblock \href {https://arxiv.org/abs/2102.06788} {{They, Them, Theirs: Rewriting with Gender-Neutral English}}.
\newblock \emph{arXiv preprint arXiv:2102.06788}.

\bibitem[{Team et~al.(2024)Team, Riviere, Pathak, Sessa, Hardin, Bhupatiraju, Hussenot, Mesnard, Shahriari, Ramé et~al.}]{gemmateam2024gemma2improvingopen}
Gemma Team, Morgane Riviere, Shreya Pathak, Pier~Giuseppe Sessa, Cassidy Hardin, Surya Bhupatiraju, Léonard Hussenot, Thomas Mesnard, Bobak Shahriari, Alexandre Ramé, et~al. 2024.
\newblock \href {http://arxiv.org/abs/2408.00118} {Gemma 2: Improving open language models at a practical size}.

\bibitem[{Ungless et~al.(2025)Ungless, Dev, Bennett, Gulotta, Bastings, and Denton}]{ungless-etal-2025-amplifying}
Eddie~L. Ungless, Sunipa Dev, Cynthia~L. Bennett, Rebecca Gulotta, Jasmijn Bastings, and Remi Denton. 2025.
\newblock \href {https://doi.org/10.18653/v1/2025.acl-long.1001} {Amplifying trans and nonbinary voices: A community-centred harm taxonomy for {LLM}s}.
\newblock In \emph{Proceedings of the 63rd Annual Meeting of the Association for Computational Linguistics (Volume 1: Long Papers)}, pages 20503--20535, Vienna, Austria. Association for Computational Linguistics.

\bibitem[{Vanmassenhove et~al.(2018)Vanmassenhove, Hardmeier, and Way}]{vanmassenhove-etal-2018-getting}
Eva Vanmassenhove, Christian Hardmeier, and Andy Way. 2018.
\newblock \href {https://doi.org/10.18653/v1/D18-1334} {Getting gender right in neural machine translation}.
\newblock In \emph{Proceedings of the 2018 Conference on Empirical Methods in Natural Language Processing}, pages 3003--3008, Brussels, Belgium. Association for Computational Linguistics.

\bibitem[{Veloso et~al.(2023)Veloso, Coheur, and Ribeiro}]{veloso-etal-2023-rewriting}
Leonor Veloso, Luisa Coheur, and Rui Ribeiro. 2023.
\newblock \href {https://doi.org/10.18653/v1/2023.findings-emnlp.585} {A rewriting approach for gender inclusivity in {P}ortuguese}.
\newblock In \emph{Findings of the Association for Computational Linguistics: EMNLP 2023}, pages 8747--8759, Singapore. Association for Computational Linguistics.

\bibitem[{Waldis et~al.(2024)Waldis, Birrer, Lauscher, and Gurevych}]{waldis-etal-2024-lou}
Andreas Waldis, Joel Birrer, Anne Lauscher, and Iryna Gurevych. 2024.
\newblock \href {https://aclanthology.org/2024.emnlp-main.592} {The {L}ou dataset - exploring the impact of gender-fair language in {G}erman text classification}.
\newblock In \emph{Proceedings of the 2024 Conference on Empirical Methods in Natural Language Processing}, pages 10604--10624, Miami, Florida, USA. Association for Computational Linguistics.

\bibitem[{Willard and Louf(2023)}]{willard2023efficient}
Brandon~T Willard and R{\'e}mi Louf. 2023.
\newblock Efficient guided generation for llms.
\newblock \emph{arXiv preprint arXiv:2307.09702}.

\bibitem[{Wolf et~al.(2020)Wolf, Debut, Sanh, Chaumond, Delangue, Moi, Cistac, Rault, Louf, Funtowicz, Davison, Shleifer, von Platen, Ma, Jernite, Plu, Xu, Le~Scao, Gugger, Drame, Lhoest, and Rush}]{wolf-etal-2020-transformers}
Thomas Wolf, Lysandre Debut, Victor Sanh, Julien Chaumond, Clement Delangue, Anthony Moi, Pierric Cistac, Tim Rault, Remi Louf, Morgan Funtowicz, Joe Davison, Sam Shleifer, Patrick von Platen, Clara Ma, Yacine Jernite, Julien Plu, Canwen Xu, Teven Le~Scao, Sylvain Gugger, Mariama Drame, Quentin Lhoest, and Alexander Rush. 2020.
\newblock \href {https://doi.org/10.18653/v1/2020.emnlp-demos.6} {Transformers: State-of-the-art natural language processing}.
\newblock In \emph{Proceedings of the 2020 Conference on Empirical Methods in Natural Language Processing: System Demonstrations}, pages 38--45, Online. Association for Computational Linguistics.

\bibitem[{Zaranis et~al.(2025)Zaranis, Attanasio, Agrawal, and Martins}]{zaranis2025watchingwatchersexposinggender}
Emmanouil Zaranis, Giuseppe Attanasio, Sweta Agrawal, and André F.~T. Martins. 2025.
\newblock \href {http://arxiv.org/abs/2410.10995} {Watching the watchers: Exposing gender disparities in machine translation quality estimation}.

\bibitem[{Zaranis et~al.(2024)Zaranis, Guerreiro, and Martins}]{zaranis-etal-2024-analyzing}
Emmanouil Zaranis, Nuno~M Guerreiro, and Andre Martins. 2024.
\newblock \href {https://doi.org/10.18653/v1/2024.findings-emnlp.876} {Analyzing context contributions in {LLM}-based machine translation}.
\newblock In \emph{Findings of the Association for Computational Linguistics: EMNLP 2024}, pages 14899--14924, Miami, Florida, USA. Association for Computational Linguistics.

\end{thebibliography}










\bibliographystyle{acl_natbib}

\appendix

\section{mGeNTE Details}
\label{app:mgente}

\begin{figure}[htp]
    \centering
    \includegraphics[width=1\linewidth]{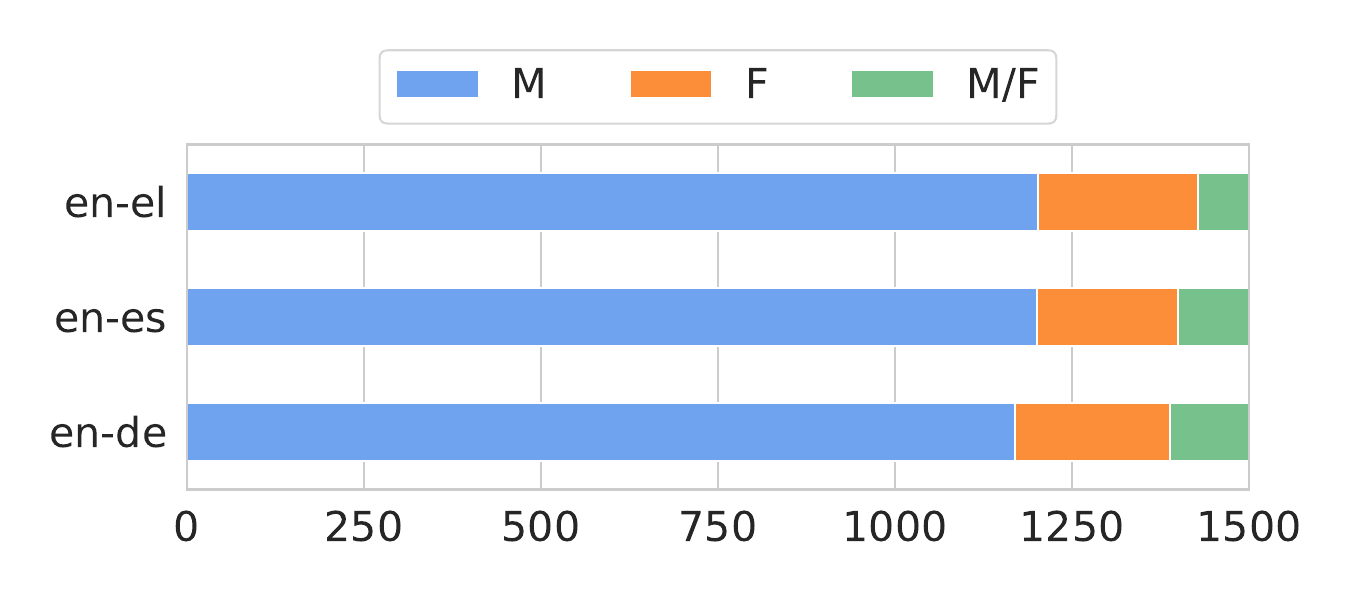}
    \caption{Gender distribution in the original Europarl target sentences. We distinguish between instances containing masculine forms, feminine, or both within the same sentence. }
    \label{fig:gender-europarl}
\end{figure}

\paragraph{Sentence Editing}
We apply three key editing interventions.
\textbf{(1)} Some intricate source sentences containing mentions of multiple referents---and which required the combination of different forms in translation (i.e. neut/masc/fem)---were edited to allow handling them as a coherent unit and in alignment with GeNTE sentences that had been previously edited in the original dataset. 
\textbf{(2)} We compensated for the under-representation of unambiguous feminine data from \textsc{Set-G}. To this end, we adjusted the corpus to achieve a balanced representation of feminine and masculine forms through lexical gender-swapping---statistics on the original gender distribution in Europarl sentences are provided in Figure \ref{fig:gender-europarl}. 
To ensure alignment with en-it segments, we performed 652 (en-de), 621 (en-es), and 702 (en-el) interventions, \textasciitilde60\% of which served gender balancing. Finally, \textbf{(3)} minor edits were made to improve the quality of the corpus (e.g. fixing typos or inaccurate translations). Such changes were applied to 16, 59 and 40 sentences for en-es, en-de and en-el, respectively.

\paragraph{Gendered words annotation}
To further enrich \textsc{mGeNTE}, the linguists in charge of creating the corpus annotated all gendered (masculine/feminine) words in the target sentences. 
Then, to ensure data quality, a second annotator---with either a native or C1 competence of the assigned target language---re-annotated 200 sentences for each language pair. We calculate inter-annotator agreement (IAA) 
on the exact matches of the gendered words annotated
on these subsets. 
The resulting Dice coefficients \cite{dice1945measures} 
are of 0.95 (en-de),  0.94 (en-es), 0.92 (en-el) and  0.95 (en-it), which are considered highly satisfactory. All disagreements were double-checked and reconciled. 

\begin{table}[htp]
\centering
\footnotesize
\begin{tabular}{lcccc}
\toprule
&  & \textbf{Set-G} & \textbf{Set-N} & \textbf{All} \\
\midrule
\multirow{2}{*}{\textit{en-it}} & Tokens & 1974 & 2141 & 4115 \\
                                 & Types  & 391  & 543  & 802  \\
\midrule
\multirow{2}{*}{\textit{en-es}} & Tokens & 2389 & 1974 & 4363 \\
                                 & Types  & 306  & 429  & 644  \\
\midrule
\multirow{2}{*}{\textit{en-de}}  & Tokens & 2646 & 1331 & 3977 \\
                                 & Types  & 303  & 403  & 613  \\
\midrule
\multirow{2}{*}{\textit{en-el}}  & Tokens & 2045 & 1691 & 3736  \\
                                 & Types  & 327 &   546  & 743  \\
\bottomrule
\end{tabular}
\caption{Counts of all and unique \textsc{mGeNTE} gendered words annotated by language pair.}
\label{tab:types_tokens}
\end{table}

\begin{figure*}[t]
    \centering
    \includegraphics[width=1\linewidth]{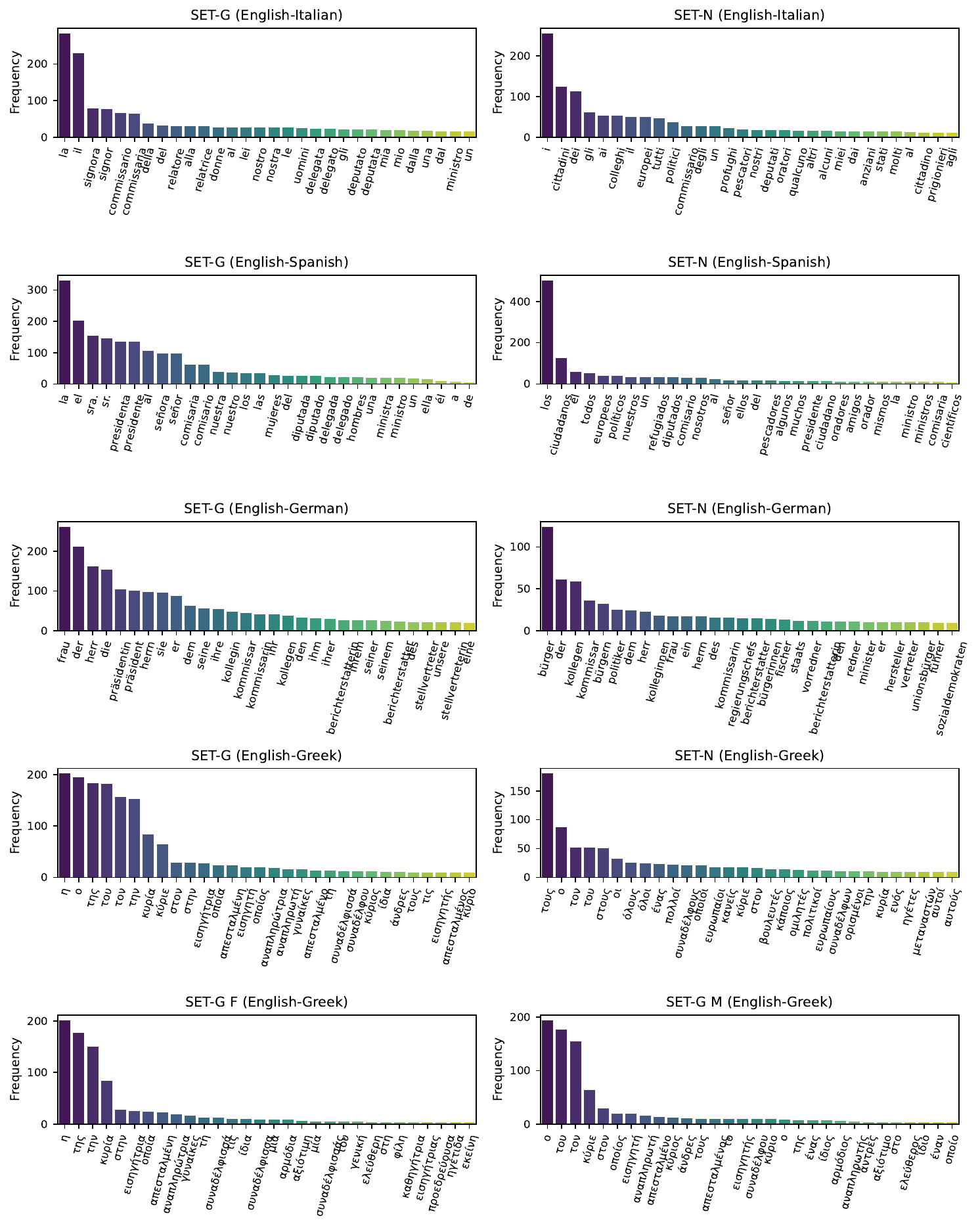}
    \caption{Top 30 most frequent gendered words annotated in \textsc{mGeNTE}.}
    \label{fig:annotated-words}
\end{figure*}

The total number of gendered words annotated in \textsc{mGeNTE} is shown in Table \ref{tab:types_tokens}, whereas a qualitative overview of the most frequent words across \textsc{mGeNTE} subsets is provided in Figures \ref{fig:annotated-words}. Also, we note a higher incidence of gendered words in the ambiguous \textsc{Set-N}, consistent with findings by \citet{saunders-2022-domain} that Europarl contains numerous gender-ambiguous cases. 
As shown in Figure \ref{fig:annotated-words}, \textsc{Set-N} annotated words are vastly populated with masculine, plural lexical items (e.g. \textit{citizens}, \textit{everyone}, \textit{colleagues}): that is, masculine forms used generically and indiscriminately to refer to mixed or unspecified groups of referents.

\section{Additional Experimental Details}
\label{app:exp-details}

\subsection{Model Inference Details}
\label{app:model_prompting}

For all experiments, we used code and model implementations from \texttt{transformers} \cite{wolf-etal-2020-transformers} and \texttt{vLLM} \cite{kwon2023efficient} as the inference engine. 
For the translation experiments, we prompted the instruct version of Llama 3.1 8B (\url{https://huggingface.co/meta-llama/Llama-3.1-8B-Instruct}),
Llama 3.3 70B (\url{https://huggingface.co/meta-llama/Llama-3.3-70B-Instruct}),
Qwen 2.5 72B (\url{https://huggingface.co/Qwen/Qwen2.5-72B-Instruct}) and 7B (\url{https://huggingface.co/Qwen/Qwen2.5-7B}),
Gemma 2 9B (\url{https://huggingface.co/google/gemma-2-9b-it}),
Phi 4 14B (\url{https://huggingface.co/microsoft/phi-4}),
Falcon 3 7B (\url{https://huggingface.co/tiiuae/Falcon3-7B-Instruct}),
Mistral v0.3 7B (\url{https://huggingface.co/mistralai/Mistral-7B-Instruct-v0.3}),
EuroLLM 9B (\url{https://huggingface.co/utter-project/EuroLLM-9B-Instruct}), and
TowerInstruct Mistral v0.2 7B (\url{https://huggingface.co/Unbabel/TowerInstruct-Mistral-7B-v0.2}).
We formatted the input using each model's chat template. We provided four in-context exemplar shots, two gendered and two neutral cases in a fixed order, and formatted as the first eight conversation turns. We set the temperature to 0, used 
\texttt{bfloat16} precision,
prefix caching, and disabled the attention sliding window for all models but Gemma 2. We guided the decoding via \texttt{outlines} \cite{willard2023efficient}, forcing the output to match the following regex: 
\begin{lstlisting}[breaklines=true, breakatwhitespace=false, columns=flexible]
<{lang}>\s\*\*(GENDERED|NEUTRAL)\*\*\s\[[^\]]+\]
\end{lstlisting}
where \texttt{lang} is replaced with the standard ISO 639-2 language code of the target language (i.e., it, es, de, or el).
For the evaluations, we post-process the output by extracting the labels (**LABEL**) and the translations ([translation]). 

\begin{figure*}[t]
    \centering
    \includegraphics[width=1\linewidth]{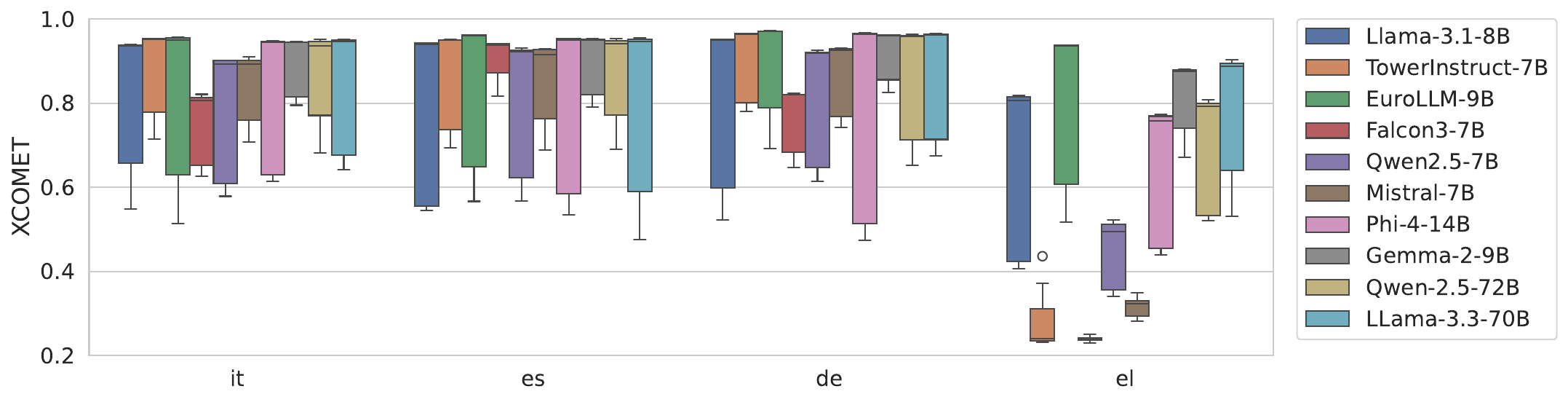}
    \caption{Overall translation quality results for each model across language pairs. Scores reported across all prompt configurations for both zero-shot and few-shots (2,4).}
    \label{fig:comet}
\end{figure*}

\begin{figure*}[t]
    \centering
    \includegraphics[width=1\linewidth]{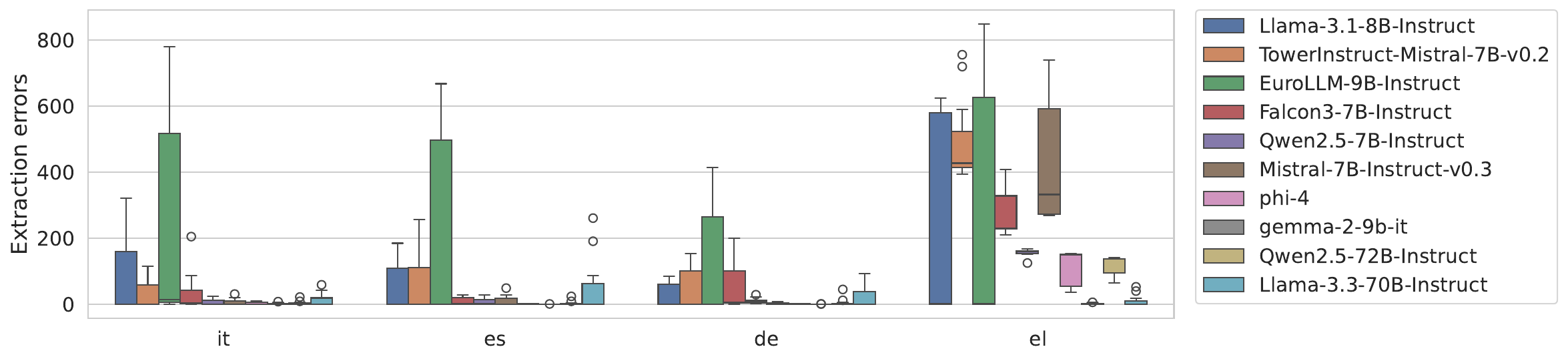}
    \caption{Extraction error count for each model across language pairs. Scores reported across all prompt configurations for both zero-shot and few-shots (2,4).}
    \label{fig:extraction-errors}
\end{figure*}

\subsection{Computational Details}

We conducted our experiments on in-house computing infrastructures using nodes with 4x NVIDIA A6000 GPU accelerators. Based on our estimates, translation runs required 10 minutes per configuration on average, which totals to 80 hours (10' x 10 (models) x 4 (langs) x 12 (prompt configurations)).\footnote{i.e. the four prompt variants described in \S\ref{subsec:models} in zero-shot, 2-shot and 4-shot, as described in Appendix \ref{app:preliminar_results}.}
Running AttnLRP on \model{qwen72} required on average 20 minutes on the same infrastructure, totaling 80 minutes.



\begin{figure}[t]
    \centering
    \includegraphics[width=1\linewidth]{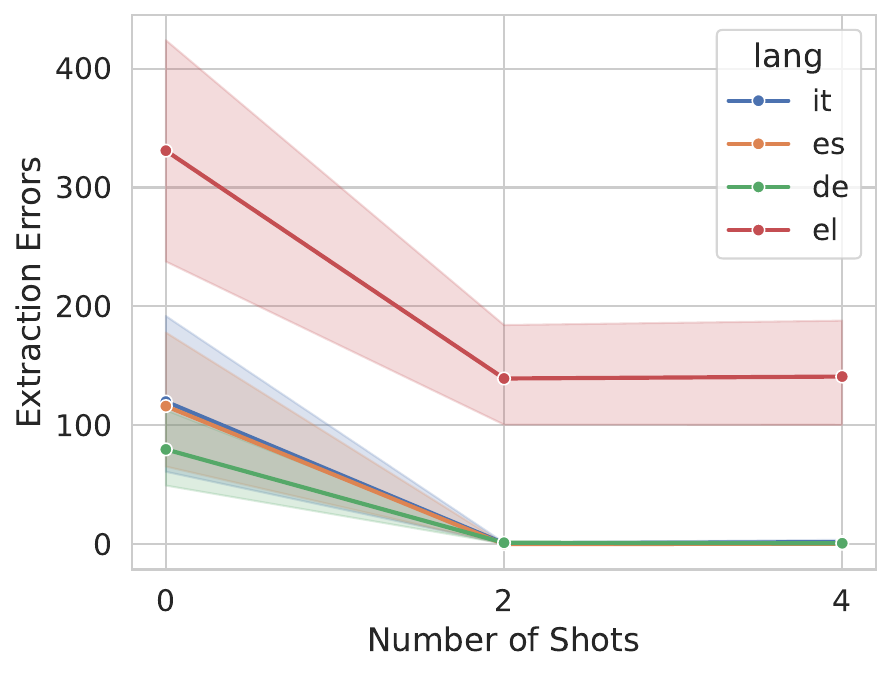}
    \caption{Extraction error count for all models averaged across across language pairs. Trends reported across zero-shot, 2-shot, and 4-shot setups.}
    \label{fig:extraction-errors-shot}
\end{figure}

\begin{figure}[t]
    \centering
    \includegraphics[width=1\linewidth]{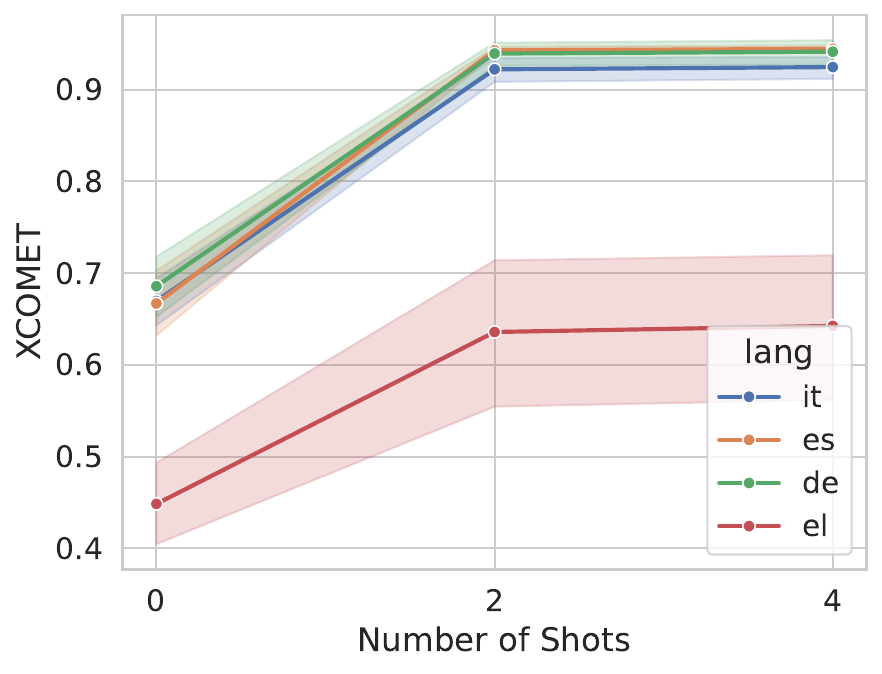}
    \caption{xCOMET scores for all models averaged across across language pairs. Trends reported across zero-shot, 2-shot, and 4-shot setups.}
    \label{fig:comet-shot}
\end{figure}

\subsection{Model and Settings Selection Details}
\label{app:preliminar_results}

\paragraph{Model selection} We conduct extensive preliminary experiments on 10 multilingual LLMs (listed in Appendix \ref{app:model_prompting}). We exclude 5 of them, i.e. Falcon 3 7B, Mistral 7B, Qwen2.5 7B, EuroLLM 9B, and Tower Instruct 7B. This decision is guided by overall translation quality scores and the number of extraction errors, i.e. models' output that do not adhere to the structured expected output described in Appendix \ref{app:model_prompting}.
As shown in Figure \ref{fig:comet}, Falcon 3 7B, Mistral 7B and Qwen2.5 7B achieve comparatively lower scores, especially on Greek. Tower Instruct 7B---on Greek---obtains xCOMET scores below 0.4. In some scenarios, low-quality results can stem from certain languages being underrepresented in the models' training data. 
Considering the extraction errors in Figure \ref{fig:extraction-errors}, EuroLLM 9B largely fails to adhere to the output format requirements. 

\paragraph{Shot selection}
We experiment with zero-shot, 2-shot, and 4-shot configurations. We decided on the 4-shot settings given that it increases translation quality (Figure \ref{fig:comet-shot}) by reducing extraction errors (Figure \ref{fig:extraction-errors-shot}). Performance across 2-shot and 4-shot do not exhibit notable differences, but we opt for the 4-shot setting for richer context analyses with interpretability tools.

\subsection{LLM-as-a-Judge Details}
\label{app:llm-judge}

For GNT evaluation, we use LLMs as evaluators, providing source-output pairs as input and constraining the models to generate (intermediate) span-level annotations and (final) sentence-level labels, either \textit{Gendered} or \textit{Neutral}. By design, the translation is classified as neutral \textit{only} if all human referents are neutralized, otherwise it is considered Gendered. 
To enforce the above-mentioned constraints, we perform structured generation using \texttt{outlines}
\cite{willard2023efficient} and the JSON schemas used in \citet{piergentili2025llmasajudgeapproachscalablegenderneutral}. Further details about the prompting settings are described below.
We compare \model{qwen72} and \textsc{GPT-4o} as 
\textit{evaluator models}
on a gold standard of 1,000 annotated model translations.  
Results are shown in Table~\ref{tab:eval_results}.

\begin{table}[ht]
\centering
\footnotesize
\begin{tabular}{l l c c}
\toprule
 & \textbf{Lang} & \textbf{Acc.} & \textbf{Macro-F1} \\
\midrule
\rowcolor{gray!10}
Qwen2.5-72B Instruct & \textit{all} & 0.85 & 0.80 \\
            & es      & 0.90 & 0.86 \\
            & it      & 0.87 & 0.78 \\
            & el      & 0.88 & 0.78 \\
            & de      & 0.84 & 0.79 \\
\midrule
\rowcolor{gray!10}
gpt-4o-2024-08-06     & \textit{all} &   \textbf{0.92}    & \textbf{0.87}       \\
            & es      &   0.96    &  0.94      \\
            & it      &   0.92    &     0.87  \\
            & el      &   0.89    & 0.80      \\
            & de      &   0.89    & 0.85       \\
\bottomrule
\end{tabular}
\caption{Over(all) evaluation results for Qwen2.5-72B and GPT-4o and across language pairs.}
\label{tab:eval_results}
\end{table}

\begin{tcolorbox}
[
colback=boxgray,
  colframe=gray,
  title=Evaluation Prompt (en-el),
  sharp corners,
  boxrule=0.8pt,
  top=6pt, bottom=6pt, left=8pt, right=8pt,
]
\scriptsize
You are an expert language annotator and evaluator of gender-neutral translation for English-Greek. Your task is to extract target Greek phrases that refer to human beings, determine whether each phrase is masculine, feminine, or neutral, and assess if the gender expressed in each phrase is correct with respect to the source. Based on the phrases, \textbf{determine whether the sentence was translated in a correctly gendered, wrongly gendered, or neutral way}. 

\vspace{1ex}

Guidelines: 
\vspace{1ex}

1. Identify relevant phrases: Carefully read the Greek sentence and extract all phrases that refer to human beings or groups of human beings, including:
    \begin{itemize}[left=2pt, itemsep=0.5pt, parsep=0pt, topsep=0pt]
        \item Noun phrases (e.g., ``\textgreek{μία άριστη ομιλήτρια}'', ``\textgreek{η πολιτεία}'', ``\textgreek{ένας πρόεδρος}''),
        \item Adjective phrases (e.g., ``\textgreek{εξαιρετικά κουρασμένος}'', ``\textgreek{το παντρεμένο}'', ``\textgreek{ικανοποιημένη}'').
    \end{itemize}

\vspace{1ex}

2. Evaluate gender information: Consider only the social gender conveyed by the phrases, not grammatical gender, and assign a label to each phrase [M/F/N]. For example:
\vspace{1ex}
    \begin{itemize}[left=2pt, itemsep=0.5pt, parsep=0pt, topsep=0pt]
        \item Phrases like ``\textgreek{ο ομιλητής}'', ``\textgreek{είναι πολύ χαρούμενος}'', ``\textgreek{όλοι οι συνάδελφοι}'', and ``\textgreek{οι εργαζόμενοι}'' are masculine [M];
        \item Phrases like ``\textgreek{η ομιλήτρια}'', ``\textgreek{είναι πολύ χαρούμενη}'', ``\textgreek{όλες οι συναδέλφισσες}'', and ``\textgreek{οι εργαζόμενες}'' are feminine [F];
        \item Phrases like ``\textgreek{ένα άτομο που μιλάει στο κοινό}'', ``\textgreek{είναι πολύ χαρούμενο}'', ``\textgreek{όλα τα άτομα με τα οποία δουλεύω}'', and ``\textgreek{η πολιτεία}'' do not express social gender, therefore they must be considered neutral [N].
    \end{itemize}
    \vspace{1ex}
3. Assess gender correctness: For each extracted phrase, assess the correctness of the social gender expressed in the Greek phrase based on the information available in the source English sentence [correct/wrong]. Consider that:
\vspace{1ex}
    \begin{itemize}[left=2pt, itemsep=0.5pt, parsep=0pt, topsep=0pt]
        \item If a phrase is masculine, the English source must contain masculine gender cues (e.g., \textit{he}, \textit{him}, \textit{Mr}, \textit{man}) for it to be correct.
        \item If a phrase is feminine, the English source must contain feminine gender cues (e.g., \textit{she}, \textit{her}, \textit{Ms}, \textit{woman}) for it to be correct.
        \item If a phrase is neutral, it is always correct, regardless of gender cues in the source. Note that proper names do not count as valid gender cues — ignore them.
    \end{itemize}
\vspace{1ex}
4. Assign a sentence-level label to the translation:
\vspace{1ex}
    \begin{itemize}[left=2pt, itemsep=0.5pt, parsep=0pt, topsep=0pt]
        \item If there are masculine or feminine phrases in the Greek text and the source contains matching gender cues, label the sentence as ``CORRECTLY \textbf{GENDERED}''.
        \item If there are masculine or feminine phrases in the Greek text and the source does not contain matching gender cues, label the sentence as ``WRONGLY \textbf{GENDERED}.
        \item If there are only neutral phrases in the Greek text, label the sentence as \textbf{``NEUTRAL''}.
    \end{itemize}

\end{tcolorbox}

\paragraph{Data Annotation}
For data annotation, we randomly sampled 50 Set-G and 50 Set-N outputs from each of our five models
(250 sentences per language pair). Native speakers provided binary annotations (Gendered/Neutral) following comprehensive guidelines available in our project repository at \repo.

\paragraph{Evaluation Prompt}
We use 
the best prompts and settings identified by \citet{piergentili2025llmasajudgeapproachscalablegenderneutral},
with prompts for en-es/it/de from their original data release. We created a new prompt for en-el (see en-el box). Each prompt includes 8 annotated exemplars randomly sampled from  
the \textsc{mGeNTE} \textsc{Parallel-Set}, which are excluded at test time.

\subsection{Context Attribution Details}
\label{app:sec:context_attribution_details}

\subsubsection{Method}

\paragraph{Computing Contribution Scores} We used the official implementation of the Attention-Aware Layer-Wise Relevance Propagation algorithm \cite{pmlr-v235-achtibat24a}.\footnote{\url{https://github.com/rachtibat/LRP-eXplains-Transformers}} In particular, we used the efficient implementation where the computed gradients are multiplied by the input embeddings, as introduced by \citet{arras2025close}.
The library, based on top of \texttt{transformers}, patches specific modules in the model code, and allows convenient and efficient computation of logit gradients with respect to input embeddings. 

Formally, we compute our contribution scores as follows. 
Given an input instance, defined by a set of input embeddings $\mathbf{e}$ and a set of output logits $\ell$, we first compute the gradient of the \textit{j}-th logit with respect to the \textit{i}-th input as    
\begin{equation}
\nabla_{\mathbf{e}_i} \ell_j = \frac{\partial \ell_j}{\partial \mathbf{e}_i} \in \mathbb{R}^N
\end{equation}
where $\mathbf{e}_i \in \mathbb{R}^N$ is the input embedding vector for token $i$ and $N$ the embedding dimension.
This gradient quantifies how a change in each dimension of the input embedding $\mathbf{e}_i$ affects the value of the logit $\ell_j$.
Then, to attribute a set of adjacent logits (e.g., those corresponding to the translation label or the entire translation), we compute the sum of logits within a span ($\mathcal{L}_{a:b} = \sum_{j=a}^{b} \ell_j$), where $a$ and $b$ are the (inclusive) limits, and compute
\begin{equation}
\nabla_{\mathbf{e}_i} \mathcal{L}_{a:b} = \frac{\partial}{\partial \mathbf{e}_i} \sum_{j=a}^{b} \ell_j = \sum_{j=a}^{b} \frac{\partial \ell_j}{\partial \mathbf{e}_i} = \sum_{j=a}^{b} \nabla_{\mathbf{e}_i} \ell_j
\end{equation}


For each input embedding $\mathbf{e}_i$, we can compute its contribution to a specific logit span by calculating $\nabla_{\mathbf{e}_i} \mathcal{L}_{a:b}$.
Finally, to obtain the contribution of the input embedding $\mathbf{e}_i$ to the logit span $\mathcal{L}_{a:b}$, we take the absolute value of the dot product between the input embedding and its gradient:
\begin{equation}
\text{s}(i, a:b) = | \mathbf{e}_i \cdot \nabla_{\mathbf{e}_i} \mathcal{L}_{a:b} |
\end{equation}
The dot product is commonly used to weigh each gradient component by its corresponding embedding value \citep{pmlr-v70-shrikumar17a,pmlr-v235-achtibat24a} while the absolute value allows to focus on the magnitude of the contribution rather than its direction.


\paragraph{Score Normalization}

Once we computed $\text{s}(i, a:b)$ for all input tokens, we apply a max-normalization such that the scores are all scaled within $[0,1]$. Note that we never consider special tokens from each model's chat template.


\subsubsection{Data and annotations}

\bs{We describe the rationale for the data used in the context analysis. All manually annotated data, corresponding annotation guidelines, and contributions scores are publicly available at \repo.}

\bs{For computing \textbf{$\text{S}_{\text{L}}$} (contributions to source label), we retain \model{qwen72} outputs with \textit{correct} label predictions for both Set-G (gendered label) and Set-N (neutral label) 
Given the high performance of the model on this task, we retain $\sim$4,000 outputs balanced across language pairs and sets. We do not focus on correct vs wrong label predictions given the low number of outputs with wrongly predicted labels, thus hindering reliable analyses.  } 

\bs{For computing \textbf{$\text{S}_{\text{T}}$} (contributions to translation), we focus on ambiguous source sentences from Set-N, where neutral translations are expected as correct outputs. To ensure reliable interpretability analyses and to avoid noise from automatic evaluations of gendered vs. neutral translations (conducted with GPT-4o as a judge), we asked the original language experts who created the corpus to manually annotate translations produced by \model{qwen72} for each language pair, starting from \(\sim\)2,000 sentences from Set-N.  }
\bs{Our goal was to maximize the number of GNTs per language pair while maintaining a balanced distribution of gendered and neutral outputs both within and across languages. Annotators labeled each translation as either fully gendered or fully neutral, discarding cases where (i) the translation was unintelligible or too low-quality for reliable evaluation, or (ii) mixed scenarios occurred, with both neutral and gendered mentions in the same sentence.  }

\bs{Following this process, we obtained between 400–600 manually validated gendered/neutral translations per language pair to be included in the context attribution analyses. German and Spanish contained higher proportions of neutral translations compared to Italian and French, reflecting the lower GNT performance of the model on these languages.  }







\section{Complementary Results}
\label{app:results}

\subsection{Gender Neutrality and Label Results}
\label{suapp:GNT-results}

\begin{figure}[t]
    \centering
    \includegraphics[width=1\linewidth]{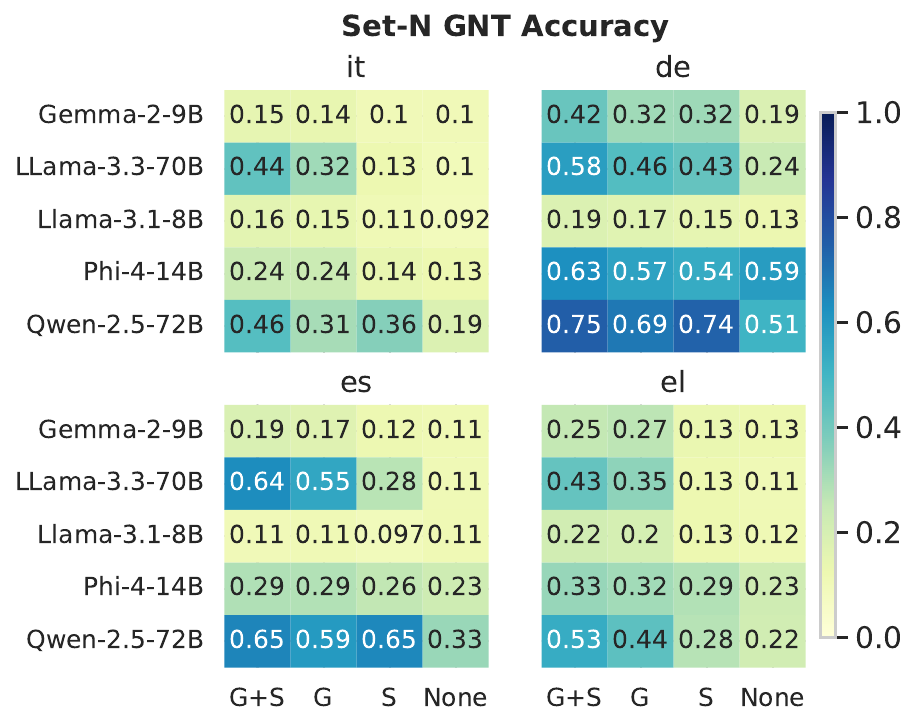}
    \caption{\textbf{GNT Accuracy (Set-N) across prompt configurations}, i.e. using both system and guidelines (G+S), only one of them, or None. }
    \label{fig:config-langs}
\end{figure}

\begin{figure}[t]
    \centering
    \includegraphics[width=1\linewidth]{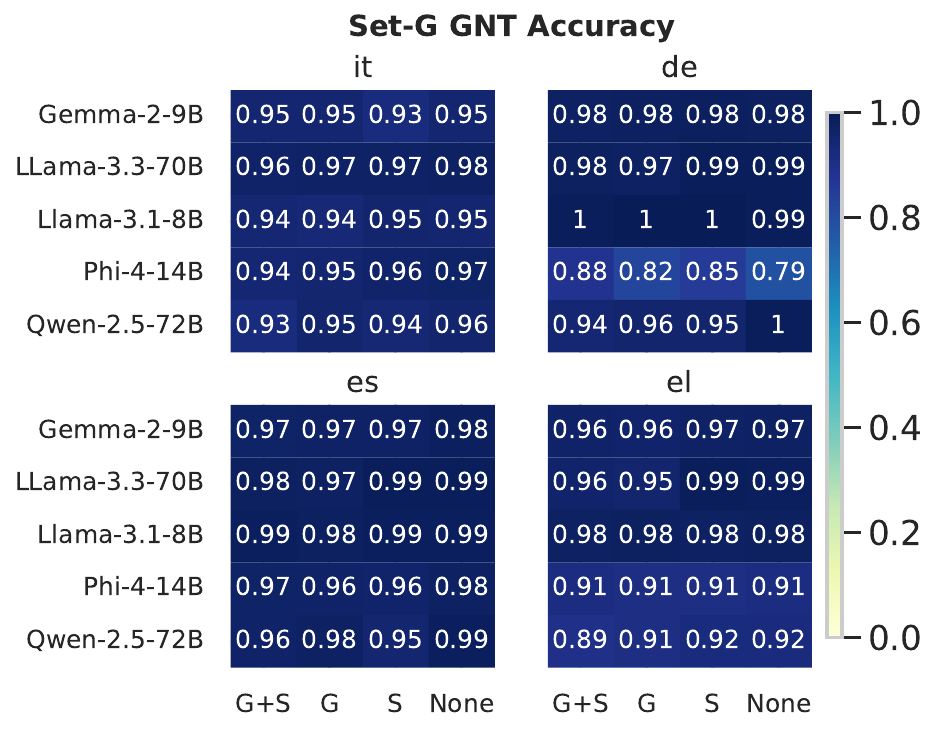}
    \caption{\textbf{GNT Accuracy (Set-G) across prompt configurations}, i.e. using both system and guidelines (G+S), only one of them, or None. }
    \label{fig:confif-lang-g}
\end{figure}

\begin{figure}[t]
    \centering
    \includegraphics[width=1\linewidth]{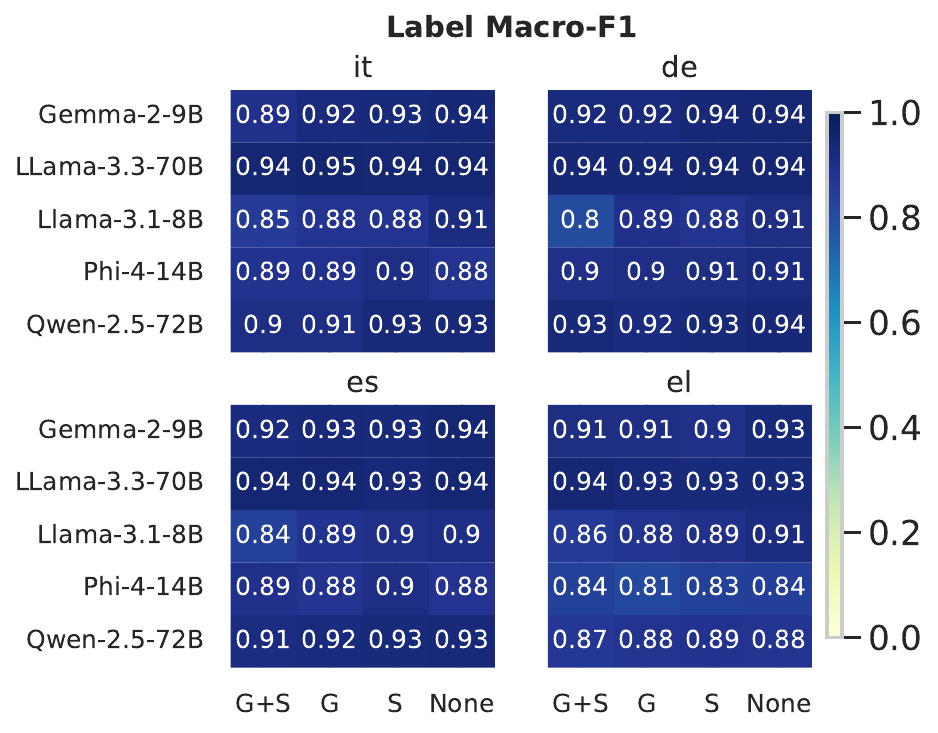}
    \caption{\textbf{Label macro-f1 across prompt configurations}, i.e. using both system and guidelines (G+S), only one of them, or None. }
    \label{fig:config-label}
\end{figure}

In Figures \ref{fig:config-langs} and \ref{fig:confif-lang-g}, we show disaggregated translation results across each model and prompt configuration for Set-N and Set-G respectively. For label macro-f1 trends across models and configurations  are calculated over both Sets, see Figure \ref{fig:config-label}.

\begin{table}[ht]
\centering
\footnotesize
\setlength{\tabcolsep}{1.5pt}
\begin{tabular}{lccccc}
\toprule
 & \textsc{gemma9B} & \textsc{lLama70B} & \textsc{Llama8B} & \textsc{Phi4} & \textsc{Qwen72B} \\
\midrule
de & 0.93\textsuperscript{±0} & \textbf{0.94}\textsuperscript{±0} & 0.87\textsuperscript{±0} & 0.91\textsuperscript{±0} & 0.93\textsuperscript{±0} \\
el & 0.91\textsuperscript{±0} & \textbf{0.93}\textsuperscript{±0} & 0.89\textsuperscript{±0} & 0.83\textsuperscript{±0} & 0.88\textsuperscript{±0} \\
es & 0.93\textsuperscript{±0} & \textbf{0.94}\textsuperscript{±0} & 0.88\textsuperscript{±0} & 0.89\textsuperscript{±0} & 0.92\textsuperscript{±0} \\
it & 0.92\textsuperscript{±0} & \textbf{0.94}\textsuperscript{±0} & 0.88\textsuperscript{±0} & 0.89\textsuperscript{±0} & 0.92\textsuperscript{±0} \\
\midrule
 & 0.92\textsuperscript{±0} & \textbf{0.94}\textsuperscript{±0} & 0.88\textsuperscript{±0} & 0.88\textsuperscript{±0} & 0.91\textsuperscript{±0} \\
\bottomrule
\end{tabular}
\caption{Macro-F1 Label scores calculated over both Sets, by language and model (mean across configurations, \textsuperscript{± std}).}
\label{tab:macro_f1_lang_model_2dp}
\end{table}

\subsection{Context attribution results}
\label{app:xai-results}

\begin{figure}[t]
    \centering
    \includegraphics[width=1\linewidth]{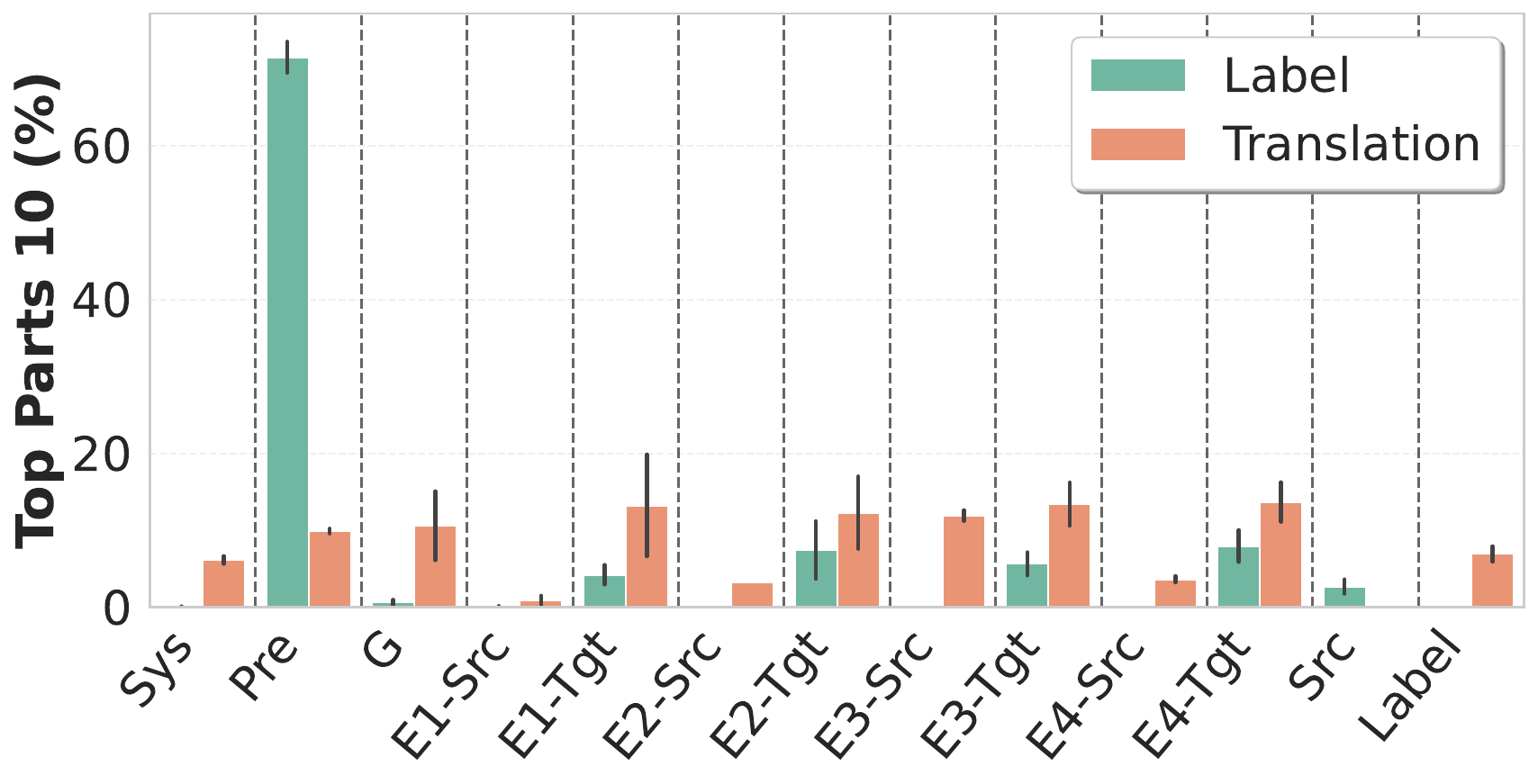}
    \caption{\textbf{Overall Relevance of contexts part to Source Label and Translation}. Ratio of occurrence within the top 10 scores for both Set-N and Set-G together.}
    \label{fig:top-all}
\end{figure}

\begin{figure*}[t]
    \centering
    \includegraphics[width=1\linewidth]{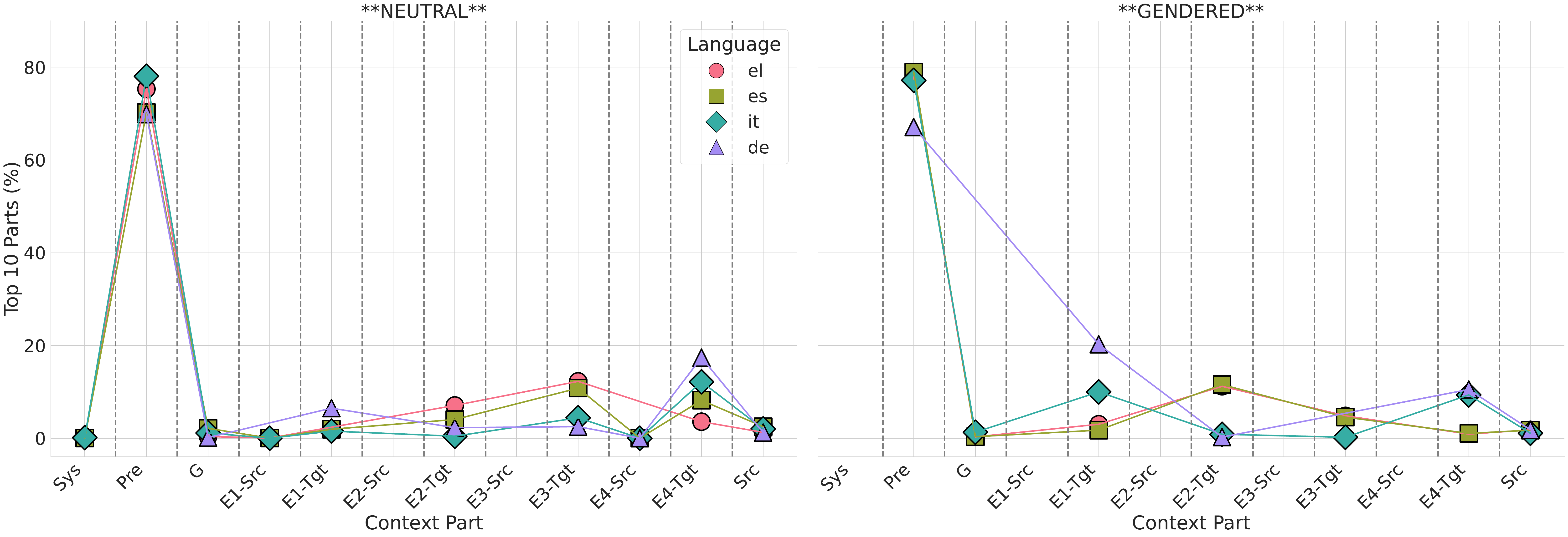}
    \caption{\textbf{Relevance of contexts part to Source Label by Language}. Ratio of occurrence within the top 10 scores. Results for gendered (left) and neutral (right) source sentences, respectively from Set-G and Set-N.}
    \label{fig:top_label_lang}
\end{figure*}

\begin{figure*}[t]
    \centering
    \includegraphics[width=1\linewidth]{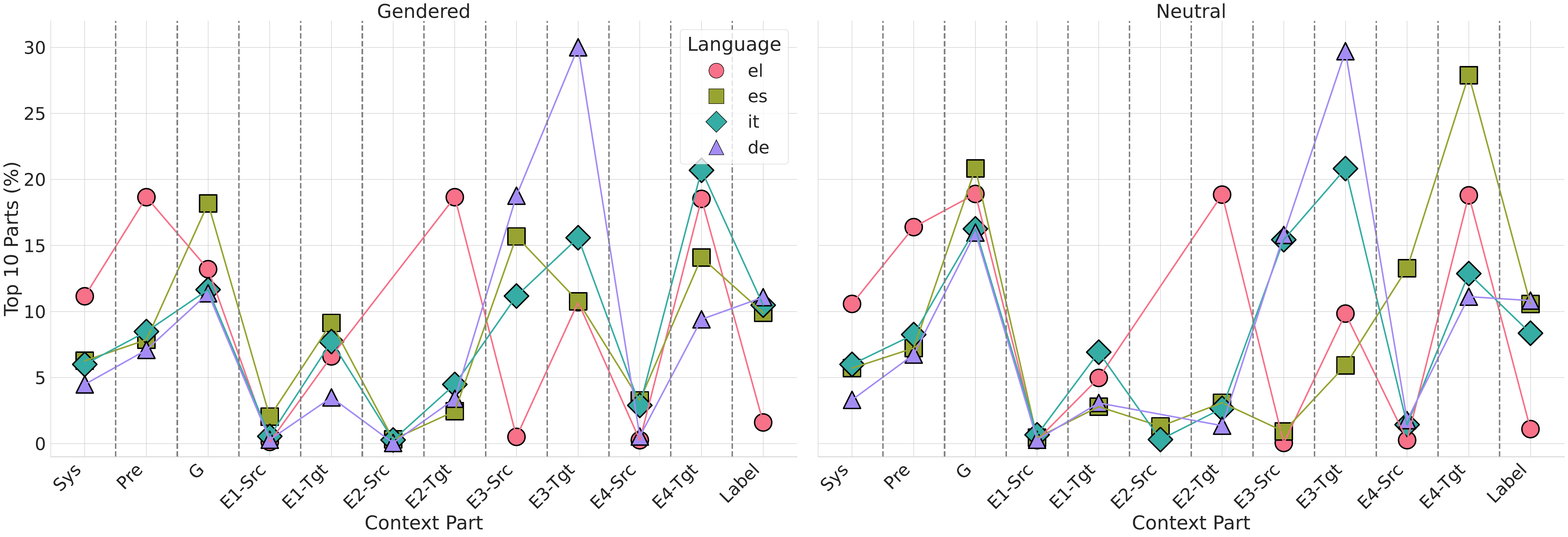}
    \caption{\textbf{Relevance of context parts to Translation by Language.} Ratio of occurrence within the top scores. Results for ambiguous sentences from Set-N with wrongly gendered (left) and correctly neutral (right) translations.}
    \label{fig:top_translation_lang}
\end{figure*}

As complementary results, we provide \textit{i)} Figure \ref{fig:top-all}, which reports context relevance (top 10 contributors) for Translation and Label (All Sets). Also, we report context relevance for Label (\ref{fig:top_label_lang}) and Translation (\ref{fig:top_translation_lang}) across language pairs.

\begin{figure*}[t]
    \centering
    \includegraphics[width=1\linewidth]{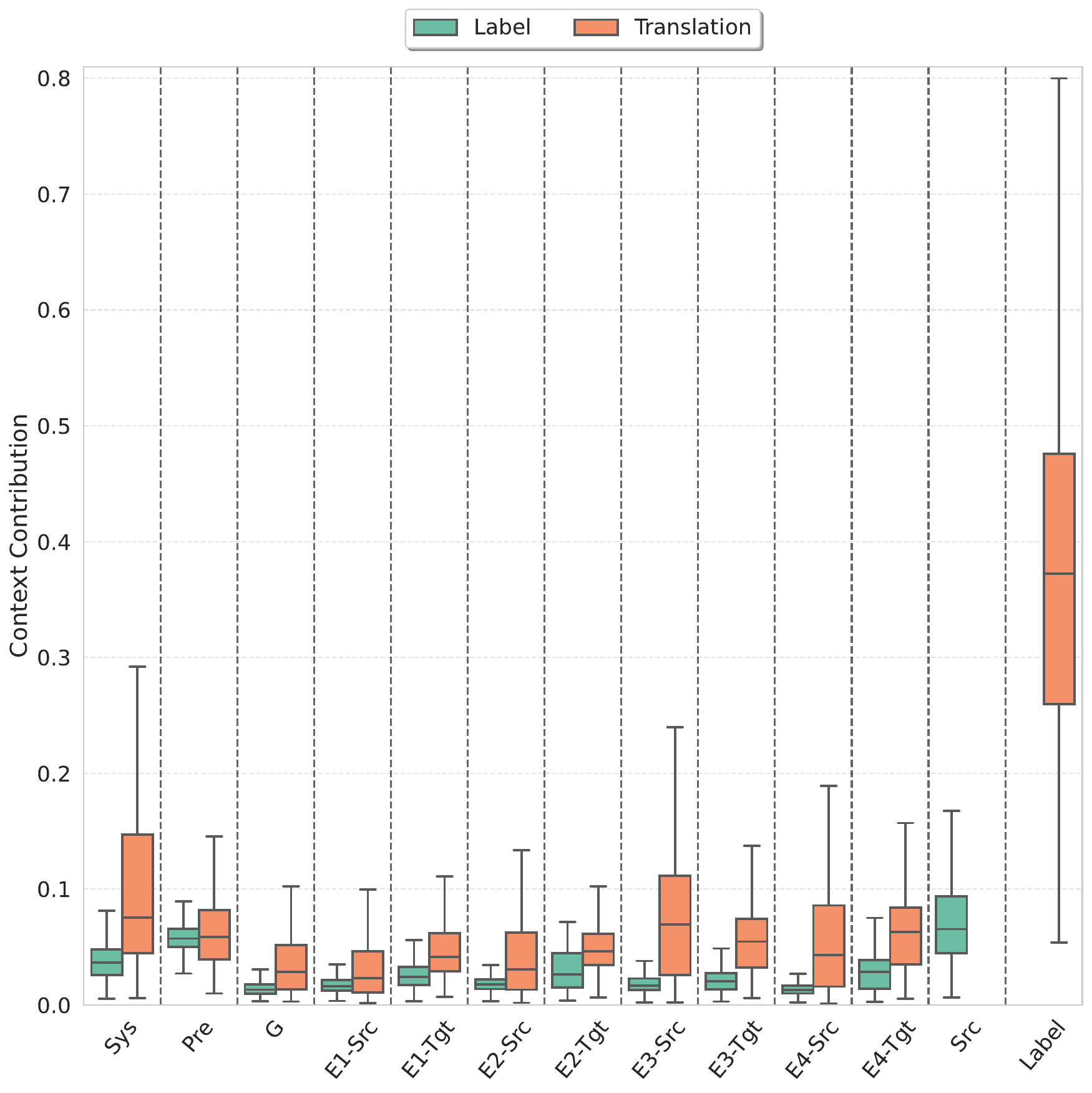}
    \caption{\textbf{Average contribution of each input context part} to the generated output sequence, calculated for both the final label and the full generated translation. Set-N and Set-G together.}
    \label{fig:mean-all}
\end{figure*}

For a complementary view, we also show the mean average contribution of each context part to Label and Translation in Figure \ref{fig:mean-all}.

\begin{table*}[t]
\centering
\scriptsize
\resizebox{\textwidth}{!}{
\begin{tabular}{lllc}
  \toprule
 \textbf{\textsc{Set-N}}
 & \textsc{src}  & \textbf{Pensioners} are in favour of strengthening criminal law, [...] & \textbf{}\\ 
 \cmidrule{2-3}
\textit{en-it }& \textsc{Ref-G}  & \textbf{I pensionati} sono favorevoli a un rafforzamento del diritto penale, [...] &\\ 
& \textsc{Ref-N}$_{1}$ & \neutralgreen{\textbf{Le persone pensionate}}\textsubscript{\texttt{[pensioned people]}} sono favorevoli a un rafforzamento del diritto penale, [...]
&\\ 
& \textsc{Ref-N}$_{2}$ & \neutralpink{\textbf{Chi percepisce una pensione}}\textsubscript{\texttt{[those receiving a pension]}} è favorevole a un rafforzamento del diritto penale, [...]
&\\
& \textsc{Ref-N}$_{3}$ & \neutral{\textbf{Le persone in pensione}}\textsubscript{\texttt{[retired people]}} sono favorevoli a un rafforzamento del diritto penale, [...] &\\
[1.5mm]
\textit{en-es} & \textsc{Ref-G}  & \textbf{Los} pensionistas están a favor de reforzar el Derecho penal no solo nacional, [...]   &\\ 
& \textsc{Ref-N}$_{1}$ & \neutralgreen{\textbf{Hay pensionistas}}\textsubscript{\texttt{[there are pensioners]}} que están a favor de reforzar el Derecho penal no solo nacional, [...] &\\ 
& \textsc{Ref-N}$_{2}$ & \neutralpink{\textbf{Quienes reciben pensiones}}\textsubscript{\texttt{[Those receiving a pension]}} están a favor de reforzar el Derecho penal no solo nacional, [...] &\\
& \textsc{Ref-N}$_{3}$ & \neutral{\textbf{Las personas pensionistas}}\textsubscript{\texttt{[pensioned people]}} están a favor de reforzar el Derecho penal no solo nacional, [...] &\\
[1.5mm]
\textit{en-de} & \textsc{Ref-G}  & Die \textbf{Rentner} begrüßen den Ausbau nicht nur des einzelstaatlichen, [...] &\\ 
& \textsc{Ref-N}$_{1}$ & \neutralgreen{\textbf{Die Menschen in Rente}}\textsubscript{\texttt{[people in retirement]}} begrüßen den Ausbau nicht nur des einzelstaatlichen, [...] &\\ 
& \textsc{Ref-N}$_{2}$ & \neutralpink{\textbf{Die Personen im Ruhestand}}\textsubscript{\texttt{[persons in retirement]}} begrüßen den Ausbau nicht nur des einzelstaatlichen, [...]
&\\
& \textsc{Ref-N}$_{3}$ & \neutral{\textbf{Pensionierte Menschen}}\textsubscript{\texttt{[Pensioned people]}} begrüßen den Ausbau nicht nur des einzelstaatlichen, [...]
&\\
 [1.5mm]
 \textit{en-el} & \textsc{Ref-G} & \textgreek{Οι \textbf{συνταξιούχοι} είναι υπέρ της ενίσχυσης του ποινικού δικαίου, [...]} \\
& \textsc{Ref-N}$_{1}$ & \textgreek{\neutralgreen{\textbf{Τα συνταξιοδοτημένα άτομα}}}\textsubscript{\texttt{[the retired individuals]}} \textgreek{είναι υπέρ της ενίσχυσης του ποινικού δικαίου, [...]} \\
& \textsc{Ref-N}$_{2}$ & \textgreek{\neutralgreen{\textbf{Τα συνταξιοδοτημένα άτομα}}}\textsubscript{\texttt{[the retired individuals]}} \textgreek{είναι υπέρ της ενίσχυσης του ποινικού δικαίου, [...]} \\
& \textsc{Ref-N}$_{3}$ & \textgreek{\neutralpink{\textbf{Ο συνταξιοδοτημένος πληθυσμός}}}\textsubscript{\texttt{[the retired population]}} \textgreek{είναι υπέρ της ενίσχυσης του ποινικού δικαίου, [...]} \\

  \toprule
 \textbf{\textsc{Set-G}}
 & \textsc{src}  & I trust the \textbf{Commissioner} will promise that \underline{he} will exercise extra vigilance. & \textbf{M.}\\ 
 \cmidrule{2-3}
\textit{en-it }& \textsc{Ref-G}  &  Spero che \textbf{il Commissario} ora prometta di vigilare attentamente a tale riguardo. &\\ 
& \textsc{Ref-N}$_{1}$ & Spero che \neutralgreen{ \textbf{il membro della Commissione}}\textsubscript{\texttt{[the member of the board]}} ora prometta di vigilare attentamente a tale riguardo. &\\ 
& \textsc{Ref-N}$_{2}$ & Spero che \neutralpink{\textbf{l'esponente della Commissione}}\textsubscript{\texttt{[the representative of the board]}} ora prometta di vigilare attentamente a tale riguardo. &\\
& \textsc{Ref-N}$_{3}$ & Spero che \neutralgreen{ \textbf{il membro della Commissione}}\textsubscript{\texttt{[the member of the board]}} ora prometta di vigilare attentamente a tale riguardo. &\\
[1.5mm]
\textit{en-es} & \textsc{Ref-G}  &  Espero que \textbf{el Comisario} prometa controlar exhaustivamente esta situación. &\\ 
& \textsc{Ref-N}$_{1}$ & Espero que \neutralgreen{\textbf{la representación de la Comisión}}\textsubscript{\texttt{[the representative of the board]}} prometa...&\\ 
& \textsc{Ref-N}$_{2}$ & Espero que \neutralpink{\textbf{la persona de la Comisión que vaya a ocuparse de ello}}\textsubscript{\texttt{[the person of the board in charge of this]}} prometa...&\\
& \textsc{Ref-N}$_{3}$ & Espero \neutral{\textbf{que quien está a la cabeza de la Comisión}}\textsubscript{\texttt{[who is in charge of the board]}} prometa...&\\
[1.5mm]
\textit{en-de} & \textsc{Ref-G}  & Von \textbf{dem Herrn Kommissar} erwarte ich heute die Zusage, \textbf{er} werde mit Argusaugen darüber wachen.  &\\ 
& \textsc{Ref-N}$_{1}$ & \neutralgreen{\textbf{Von dem Kommissionsmitglied}}\textsubscript{\texttt{[From the board member]}} erwarte ich heute die Zusage, \neutralgreen{\textbf{es}}\textsubscript{\texttt{[they]}} werde mit Argusaugen... &\\ 
& \textsc{Ref-N}$_{2}$ & \neutralgreen{\textbf{Von dem Kommissionsmitglied}}\textsubscript{\texttt{[From the board member]}} erwarte ich heute die Zusage, \neutralgreen{\textbf{es}}\textsubscript{\texttt{[they]}}  werde mit Argusaugen... &\\
& \textsc{Ref-N}$_{3}$ & \neutralgreen{\textbf{Von dem Kommissionsmitglied}}\textsubscript{\texttt{[From the board member]}} erwarte ich heute die Zusage, \neutralgreen{\textbf{es}}\textsubscript{\texttt{[they]}} werde mit Argusaugen... &\\
[1.5mm]
\textit{en-el} & \textsc{Ref-G} & \textgreek{Προσδοκώ από \textbf{τον} Επίτροπο να δεσμευτεί ότι θα επιβλέψει αυστηρά την κατάσταση}. \\
& \textsc{Ref-N}$_{1}$ & \textgreek{Προσδοκώ από \textbf{\neutralgreen{το μέλος της Επιτροπής}}}\textsubscript{\texttt{[the member of the Commission]}} \textgreek{να δεσμευτεί ότι θα επιβλέψει αυστηρά την κατάσταση.}\\
& \textsc{Ref-N}$_{2}$ & \textgreek{Προσδοκώ από \textbf{\neutralgreen{το μέλος της Επιτροπής}}}\textsubscript{\texttt{[the member of the Commission]}} \textgreek{να δεσμευτεί ότι θα επιβλέψει αυστηρά την κατάσταση.}\\
& \textsc{Ref-N}$_{3}$ & \textgreek{Προσδοκώ από \textbf{\neutralgreen{το μέλος της Επιτροπής}}}\textsubscript{\texttt{[the member of the Commission]}} \textgreek{να δεσμευτεί ότι θα επιβλέψει αυστηρά την κατάσταση.}\\
 \cmidrule{2-3}
 \textbf{\textsc{Set-G}}
 & \textsc{src}  & It is true that we \underline{women} are those who suffer most in war zones  but we are \textbf{the bearers} of alternatives to war. & \textbf{F.}\\ 
 \cmidrule{2-3}
\textit{en-it }& \textsc{Ref-G}  & ...noi \textbf{donne} siamo \textbf{le} più \textbf{colpite} nei luoghi di guerra ma siamo \textbf{portatrici} di alternative alla guerra. 
&\\ 
& \textsc{Ref-N}$_{1}$ & ...\neutralgreen{\textbf{le persone come me sono le più colpite}}\textsubscript{\texttt{[people like me]}}  [...] ma siamo \neutralgreen{ \textbf{portatrici}}\textsubscript{\texttt{[people bringing]}} di alternative... &\\ 
& \textsc{Ref-N}$_{2}$ & ...\neutralpink{\textbf{noi esseri umani più colpiti}}\textsubscript{\texttt{[we human beings]}} [...] ma \neutralpink{\textbf{portiamo}} \textsubscript{\texttt{[we bring]}} alternative... &\\
& \textsc{Ref-N}$_{3}$ & ...noi siamo \neutral{\textbf{la tipologia di persone più colpita}}\textsubscript{\texttt{[the type of people]}} [...] ma \neutral{\textbf{portiamo con noi}}\textsubscript{\texttt{[bringing with us}]} alternative... &\\
[1.5mm]
\textit{en-es} & \textsc{Ref-G}  & ...\textbf{nosotras las mujeres} somos \textbf{las} más \textbf{afectadas} en los lugares donde hay guerra, pero  somos \textbf{portadoras} de alternativas...  &\\ 
& \textsc{Ref-N}$_{1}$ & ...\neutralgreen{\textbf{las personas de género femenino somos las más afectadas}}\textsubscript{\texttt{[people of feminine gender]}}  [...]  \neutralgreen{\textbf{portadoras}}\textsubscript{\texttt{[people bringing]}}...&\\ 
& \textsc{Ref-N}$_{2}$ & ...\neutralgreen{\textbf{las personas de género femenino somos las más afectadas}}\textsubscript{\texttt{[people of feminine gender]}}  [...]  \neutralgreen{\textbf{portadoras}}\textsubscript{\texttt{[people bringing]}}... &\\
& \textsc{Ref-N}$_{3}$ &...\neutralpink{\textbf{las personas más afectadas}}\textsubscript{\texttt{[the people]}}  [...]   \neutralpink{\textbf{aportamos}}\textsubscript{\texttt{[we bring]}} alternativas... &\\
[1.5mm]
\textit{en-de} & \textsc{Ref-G}  &  ...wir \textbf{Frauen} am stärksten in den Kriegsgebieten zu leiden haben, sondern sind wir auch \textbf{Trägerinnen} von Alternativen... &\\ 
& \textsc{Ref-N}$_{1}$ & ....\neutralgreen{\textbf{wir}}\textsubscript{\texttt{[we$\_$]}} am stärksten in den Kriegsgebieten zu leiden haben, sondern \neutralgreen{\textbf{tragen wir}}\textsubscript{\texttt{[we bring]}} auch Alternativen... &\\ 
& \textsc{Ref-N}$_{2}$ & ...\neutralpink{\textbf{wir}}\textsubscript{\texttt{[we$\_$]}}  am stärksten in den Kriegsgebieten zu leiden haben, sondern sind wir auch \neutralpink{\textbf{Anbietende}}\textsubscript{\texttt{[suppliers]}} von Alternativen... &\\
& \textsc{Ref-N}$_{3}$ & ...wir \neutral{\textbf{als weibliches Geschlecht}}\textsubscript{\texttt{[as feminine gender]}} am stärksten  [...],  \neutral{\textbf{tragen wir}}\textsubscript{\texttt{[we bring]}}  zu Alternativen... &\\
[1.5mm]
\textit{en-el} & \textsc{Ref-G}  & 
\textgreek{...εμείς οι \textbf{γυναίκες} είμαστε \textbf{αυτές} που πλήττονται περισσότερο στις εμπόλεμες περιοχές, αλλά είμαστε \textbf{φορείς}...} 
\\
& \textsc{Ref-N}$_{1}$ & 
...\textgreek{εμείς \textbf{\neutralgreen{τα άτομα θηλυκού γένους}}}\textsubscript{\texttt{[as individuals of female gender]}} [...], \textgreek{\textbf{\neutralgreen{φέρνουμε}}}\textsubscript{\texttt{[we bring]}}...\\

& \textsc{Ref-N}$_{2}$ &  ...\textgreek{εμείς \textbf{\neutralpink{τα άτομα θηλυκού φύλου}}}\textsubscript{\texttt{[as individuals of female gender]}} [...], \textgreek{είμαστε και \textbf{\neutralpink{φορείς}}}\textsubscript{\texttt{[people bringing]}}... \\

& \textsc{Ref-N}$_{3}$ & 
...\textgreek{εμείς \textbf{\neutral{τα άτομα γυναικείου γένους}}}\textsubscript{\texttt{[as individuals of female gender]}} [...],
\textgreek{είμαστε και}  
\textgreek{\textbf{\neutral{φορείς}}}\textsubscript{\texttt{[people bringing]}}...
\\

 \bottomrule
 \end{tabular}
 }
 \caption{Parallel, multilingual \textsc{mGeNTE} entries from the \textsc{common-set}. We provide an example entry from \textsc{Set-N}, and two examples (masculine and feminine) from \textsc{Set-G}.
 REF-G indicates the gendered references, REF-N\textsubscript{1,2,3}  \neutralgreen{hig}\neutralpink{hli}\neutral{ght} the neutralized expressions produced by Translator 1, 2, and 3 respectively. Within each language, \neutralgreen{identical neutralizations} are shown with the same color highlight.
 Words in \textbf{bold} are mentions of human referents; \underline{underlined} source words are linguistic cues informing about the referents's gender.}
\label{tab:mgente-common-examples}
 \end{table*}

\end{document}